\documentclass{article}

\usepackage[margin=1in]{geometry}

\usepackage{mathpazo}
\usepackage[numbers]{natbib}

\usepackage[utf8]{inputenc} 
\usepackage[T1]{fontenc}    
\usepackage[colorlinks]{hyperref}       
\usepackage{url}            
\usepackage{booktabs}       
\usepackage{amsfonts}       
\usepackage{nicefrac}       
\usepackage{microtype}      
\usepackage[table]{xcolor}         

\usepackage{color}
\usepackage{caption}
\usepackage{makecell}
\usepackage{wrapfig}
\usepackage{textcomp}
\usepackage{gensymb}
\usepackage{marginnote}
\usepackage{algorithm}
\usepackage{algorithmic}
\usepackage{enumitem}

\usepackage{tcolorbox}

\definecolor{Gray}{gray}{0.9}
\definecolor{BrickRed}{rgb}{0.6,0,0}
\definecolor{RoyalBlue}{rgb}{0,0,0.8}
\definecolor{Tdgreen}{rgb}{0,0.4,0.7}
\hypersetup{citecolor=BrickRed, linkcolor=RoyalBlue}

\usepackage{bm}
\usepackage{dsfont}
\usepackage{multirow}
\usepackage{multicol}
\usepackage{array}
\usepackage{bbding}
\usepackage{lipsum}
\usepackage{verbatim}

\usepackage{pifont}

\newcommand{\keywords}[1]{\par\textbf{Keywords:} #1}

\usepackage{microtype}
\usepackage{graphicx}
\usepackage{subcaption}
\usepackage{wrapfig}
\usepackage{booktabs} 
\usepackage{arydshln}

\usepackage{amsmath}
\usepackage{amssymb}
\usepackage{mathtools}
\usepackage{amsthm}

\usepackage[capitalize,noabbrev]{cleveref}

\theoremstyle{plain}
\newtheorem{theorem}{Theorem}[section]

\theoremstyle{definition}

\theoremstyle{remark}

\title{Unified Classification and Rejection: A One-versus-All Framework}

\author{
Zhen Cheng\textsuperscript{\rm 1,2}, Xu-Yao Zhang\textsuperscript{\rm 1,2}, Cheng-Lin Liu\textsuperscript{\rm 1,2}\\
\\
\normalsize \textsuperscript{\rm 1} MAIS, Institute of Automation, Chinese Academy of Sciences, Beijing 100190, China\\
\normalsize \textsuperscript{\rm 2} School of Artificial Intelligence, University of Chinese Academy of Sciences, Beijing, 100049, China\\
}
\date{}

\begin{document}

\maketitle

\begin{abstract}
	Classifying patterns of known classes and rejecting ambiguous and novel (also called as out-of-distribution (OOD)) inputs are involved in open world pattern recognition. Deep neural network models usually excel in closed-set classification while performs poorly in rejecting OOD inputs. To tackle this problem, numerous methods have been designed to perform open set recognition (OSR) or OOD rejection/detection tasks. Previous methods mostly take post-training score transformation or hybrid models to ensure low scores on OOD inputs while separating known classes. In this paper, we attempt to build a unified framework for building open set classifiers for both classification and OOD rejection. We formulate the open set recognition of $ K $-known-class as a $ (K+1) $-class classification problem with model trained on known-class samples only. By decomposing the $ K $-class problem into $ K $ one-versus-all (OVA) binary classification tasks and binding some parameters, we show that combining the scores of OVA classifiers can give $ (K+1) $-class posterior probabilities, which enables classification and OOD rejection in a unified framework. To maintain the closed-set classification accuracy of the OVA trained classifier, we propose a hybrid training strategy combining OVA loss and multi-class cross-entropy loss. We implement the OVA framework and hybrid training strategy on the recently proposed convolutional prototype network and prototype classifier on vision transformer (ViT) backbone. Experiments on popular OSR and OOD detection datasets demonstrate that the proposed framework, using a single multi-class classifier, yields competitive performance in closed-set classification, OOD detection, and misclassification detection. The code is available at \url{https://github.com/zhen-cheng121/CPN_OVA_unified}.
\end{abstract}

\keywords{Open set recognition, out-of-disctribution detection, misclassification detection, convolutional prototype network, one-versus-all, Dempster–Shafer theory of evicence}

\section{Introduction}
\label{sec:sec1}

In recent years, deep neural networks (DNNs) have been popularly used in pattern recognition applications because of their excellent classification performance resulted from feature representation learning, unlike traditional classifiers that rely on human design of feature extraction. The high classification performance is desired in high-stake applications such as person identification~\cite{zheng_2015_scalable}, medical diagnosis~\cite{esteva2017dermatologist}, and autonomous driving~\cite{bojarski2016end}. Besides the classification accuracy, however, DNNs are found to be vulnerable to abnormal or outlier inputs (also called as anomaly, novelty, unknown, out-of-distribution (OOD), etc.), \textit{e.g.}, samples that are out of the known classes (those have samples in training the classifier). Since the primary goal of DNN training is to separate the known classes in feature space, it results in the division of the feature space into decision regions corresponding to the known classes. So, the outlier inputs are likely to fall in the decision region of a known class and be classified to the known class with high confidence, bringing the so-called over-confidence problem~\cite{Nguyen_2015_DeepNN}. To tackle this problem, researchers have proposed many methods to perform so-called open set recognition (OSR)~\cite{bendale2016towards} or OOD rejection/detection tasks~\cite{hendrycks_2017_baseline,salehi_2021_OODsurvey}.

OOD detection could be viewed as a sub-task of OSR. It concerns mainly the differentiation of OOD samples from known-class (also called as in-distribution (InD)) samples. Because of the critical inferiority of DNNs to OOD data and the needs in real-world applications, OOD detection or rejection has received tremendous attention in recent years. OOD detection, aimed at distinguishing the OOD inputs from InD inputs, can be regarded as a binary classification task. Since the output scores of a classifier trained with a multi-class classification objective (such as cross-entropy (CE)) cannot separate OOD inputs well, one line of approaches designs new scoring functions to enlarge the gap of outputs between InD and OOD examples. A well-known baseline called Maximum Softmax Probability (MSP)~\cite{hendrycks_2017_baseline} uses the output of the softmax function as the score for OOD detection. Some newly proposed methods have been shown to achieve significant improvement in OOD detection, such as energy score~\cite{liu_2020_energy}, ViM~\cite{wang2022vim}, and KNN~\cite{sun2022KNN}. Another line of methods utilizes auxiliary outlier data since the vulnerability of models on OOD inputs is due to the lack of knowledge on the distribution of unknown classes. One of the most effective methods called Outlier Exposure (OE)~\cite{hendrycks_2019_oe}, is to train the models with auxiliary data of natural outlier images. However, outlier data is not always available in training and it is hard to guarantee that the outlier data covers various pattern styles. So, to design classifiers without outlier data training is preferred.

\begin{figure*}[t]
	\begin{center}
		\includegraphics[width=0.90\linewidth]{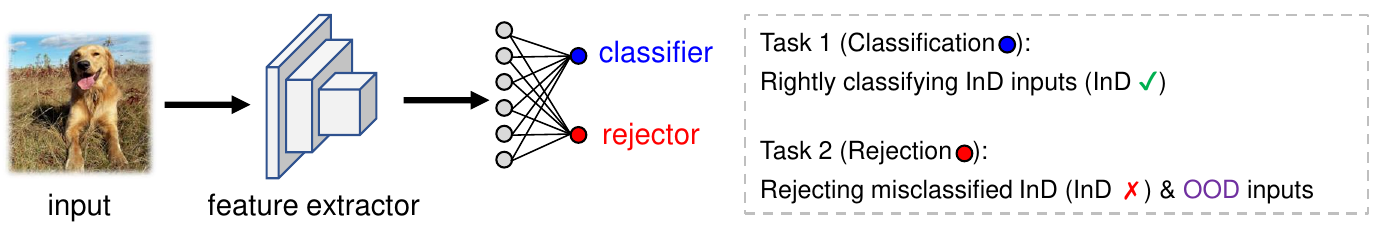}
	\end{center}
	\caption{A model \textbf{unifying classification and rejection}. The model's primary objective is to classify InD samples as accurately as possible (InD \textcolor{green}{\CheckmarkBold}), and also to reject possibly misclassified samples (InD \textcolor{red}{\XSolidBrush}) and \textcolor{purple!70}{OOD} samples.}
	\label{fig:task_framework}
	\vskip -0.10in
\end{figure*}

Open set recognition, on the other hand, considers both InD classification and OOD rejection/detection, and assumes no OOD data is available in training. This is also called as a generalized OOD detection problem~\cite{yang_2021_oodsurvey}. The OOD detection methods can be applied to or integrated into OSR, and meanwhile, the separation of known classes is considered. The numerous methods of OSR can be categorized in the dimension of model architecture (generative, discriminative, hybrid) or learning/scoring strategy (end-to-end learning or post-scoring). Popular DNNs such as convolutional neural networks (CNNs) are discriminative models, which are trained to optimize a classification objective such as multi-class CE, which leads to the separation of known classes in the feature space, disregarding the boundary of OOD inputs. Generative models, such as auto-encoders (AE)~\cite{yoshihashi_2019_CROSR,sun_2020_conditional}, learn a subspace or local region for each class, such that OOD detection can be based on the density function~\cite{sun_2020_conditional} or the distance to the local region (or prototype)~\cite{lee_2018_maha} of known classes. Hybrid methods such as~\cite{zhang_2020_hybrid} use multiple models for handling classification and rejection tasks separately. In the dimension of learning, the methods CPN~\cite{yang_2018_CPN,yang_2022_CPN} and class-specific semantic reconstruction (CSSR)~\cite{huang_2022_CSSR} are end-to-end trained models, which take into account the local distribution of known classes so that the output distance scores can differentiate OOD inputs. Many other methods~\cite{hendrycks2022MaxL} design OOD scoring functions based on the logit outputs of a pre-trained network which did not consider OOD boundary in training.

In this paper, we attempt to build a unified framework for OSR performing InD classification and OOD using a single classifier without the need for OOD data during training, as shown in Fig.~\ref{fig:task_framework}. To model the boundary between InD and OOD classes, we take advantage of one-versus-all (OVA) learning, which treats one known class as positive while the union of the rest known-class samples as negative, so the OOD class is included in the negative class, and learning can proceed no matter whether OOD samples are available or not. If we further assume a local region distribution for each known class, using a density function, subspace, or prototypes, OOD samples can be identified according to maximum density or minimum distance to known classes. The parameters of the local regions are learnable in OVA learning. We show that the OVA classifier can be analyzed in the framework of Bayesian classification, by combining the binary posterior probabilities into multi-class probabilities. For a $ K $-class classification problem, combining the OVA probabilities using the Dempster-Shafer theory of evidence (DSTE)~\cite{shafer_1976_mathematical,barnett_1981_computational}, we can obtain the posterior probabilities of $ K+1 $ classes (including the OOD class). This result enables the classification of known classes, the rejection of OOD and ambiguity (misclassification) in a unified framework based on the Bayes decision rule (maximum posterior probability decision). Considering that OVA learning may sacrifice the InD classification accuracy because of the independent training of class logits disregarding the boundary between multiple classes, we propose a hybrid learning strategy by combining OVA and multi-class objective (such as CE), so as to maintain closed-set classification accuracy while achieve OOD rejection. 

To validate the proposed unified OSR framework, we implement the OVA and hybrid learning strategy with the recently proposed convolutional prototype network (CPN)~\cite{yang_2018_CPN} and prototype classifier with vision transformer (ViT) backbone~\cite{dosovitskiy_2021_ViT}. Our experiments on popular OSR and OOD datasets verify that the proposed framework yields competitive performance in closed-set classification, OOD detection, and misclassification detection using a single classifier without training with OOD samples. 

We would like to discuss more about the \textit{\textbf{importance of a unified framework}}: (1)~A unified model is more suitable for real scenarios. Taking the task of disease diagnosis as an example, the diagnosis system may simultaneously meet examples from unseen class and misclassified examples from seen class. In addition to the ability of classification, a reliable recognizer should be able to detect both of these abnormal conditions simultaneously. (2)~Proposing a unified framework is a challenging problem. The recent study~\cite{jaeger_2023_call} has raised a call to deal with OOD detection and MisD simultaneously, and it reveals that the simultaneous detection of two types of failures is challenging. In this paper, we aim to deal with this challenge one step forward.

Our contributions in the paper can be summarized as follows:
\begin{itemize}
	\item We propose a unified framework for OSR, performing multi-class classification and OOD detection using a single classifier, trained by OVA learning without OOD samples.
	\item We show that by combining the binary posterior probabilities of OVA classifiers using the DSTE, $ (K+1) $-class posterior probabilities can be obtained from the OVA classifier outputs of $ K $ known classes, thus enabling classification and rejection in the Bayesian classification framework.
	\item To maintain the closed-set classification accuracy of OVA classifier, we propose a hybrid learning strategy, by combining OVA and multi-class CE objectives to guarantee both classification and rejection performance.
	\item We implement the OVA framework and hybrid learning strategy with the convolutional prototype network and ViT backbone and verify its superior performance in classification, OOD and misclassification detection on popular OSR and OOD datasets.
\end{itemize}

\noindent The rest of this paper is organized as follows. Section~\ref{sec:sec2} reviews related works;
Section~\ref{sec:sec3} introduces the unified OVA framework for unifying classification and rejection; Section~\ref{sec:sec4} proposes the hybrid learning strategy for training OVA classifier; Section~\ref{sec:sec5} describes the instantiation of the OVA framework with the CPN; Section~\ref{sec:sec6} presents experiments of classification and rejection on multiple datasets;
Section~\ref{sec:sec7} provides the conclusion and future works.

\section{Related Work}
\label{sec:sec2}

\noindent\textbf{Open Set Recognition.} Various methods have been proposed to trackle the task of openset recognition based on deep neural networks~\cite{geng_2021_OSRsurvey,salehi_2021_OODsurvey,yang_2021_oodsurvey}, which could be roughly categorized to AE-based methods and prototype-based methods. An earlier work of Bendale \textit{et al.}~\cite{bendale2016towards} proposed to use OpenMax to replace the softmax function in DNNs. To better utilize the information about reconstruction for OSR, Yoshihashi \textit{et al.} proposed CROSR~\cite{yoshihashi_2019_CROSR}, which employed an encoder network to produce better latent features for better reconstruction task, following C2AE~\cite{oza_2019_c2ae}. Zhang \textit{et al.}~\cite{zhang_2020_hybrid} later used normalized flow to attain the latent feature, rather than AE. Besides these methods, Yang \textit{et al.}~\cite{yang_2018_CPN} and Chen \textit{et al.}~\cite{chen_2020_RPL} proposed to use prototypes to constrain the closed region of known classes, and have achieved significant improvement on OSR.

\noindent\textbf{OOD detection.} OOD detection~\cite{hendrycks_2017_baseline}, considering primarily the separation of OOD inputs from known classes,  has received significant attention recently, and various methods have been proposed under the setting of whether to use the auxiliary data of natural outliers or not. The work of~\cite{hendrycks_2017_baseline} first introduced a baseline called maximum softmax probability (MSP) as the scoring function to detect OOD inputs. Later, various methods focused on designing scoring functions have been proposed, like ODIN~\cite{liang_2018_ODIN}, Mahalanobis distance~\cite{lee_2018_maha}, energy score~\cite{liu_2020_energy}, Gram matrix based score~\cite{Sastry2020Gram}, ViM~\cite{wang2022vim}, and KNN~\cite{sun2022KNN}. The methods mentioned above are based on post-hoc techniques. There are also many training time methods like LogitNorm~\cite{wei2022logitnorm}. AoP~\cite{cheng_2023_AoP} delved into training dynamics of OOD detection and found it suffered from instability and overfitting. OE~\cite{hendrycks_2019_oe} first introduced an extra dataset consisting of outliers in training, and enforced low confidence on such outliers. To better utilize the auxiliary dataset, different methods are proposed~\cite {Ming2022poem,zhu_2023_openmix}. Besides, to alleviate the cost of collecting outliers manually, there are some methods to synthesize outliers, like VOS~\cite{du2022vos}. 

\noindent \textbf{Rejection rules.} The performance of model rejection highly relies on a well-designed rejection rule~\cite{liu_2002_performance}. Chow~\cite{chow_1970_optimum} first proved that Bayes classifier (outputting posterior probabilities) can reject wrongly classified examples (ambiguity rejection, or called misclassification detection (MisD)) optimally, \textit{i.e.}, minimizing misclassification and rejection risks. However, considering that the Bayes optimal classifier is hard to attain in practical scenarios and actual classifier models approximate the Bayes classifier in different ways, various surrogate rules have been proposed, such as the rule based on the difference of top two discriminant functions (logits)~\cite{liu_2002_performance}. In OOD detection, the rejection rules try to approximate the marginal distribution of InD data, such as the energy score~\cite{liu_2020_energy} and some methods based on distance~\cite{lee_2018_maha,sun2022KNN}. The recent progress in rejection can be found in the survey~\cite{Zhang_2023_SurveyReject}. It is noteworthy that similar tasks of OOD detection and MisD have been studied in early works of pattern recognition, which are called distance rejection and ambiguity rejection, respectively~\cite{dubuisson_1993_statistical}.

\noindent\textbf{One-versus-all learning.} OVA training strategy has been used in machine learning for a long period~\cite{galar_2011_ova}, which regards multi-class classification as a combination of binary classifiers. The OVA training strategy is also used to reject unknown class inputs. Nevertheless, it is primarily used to combine standard CNNs with linear classifiers. Saito et al.~\cite{saito_2021_ovanet} uses OVA loss to enhance the ability to reject unknown classes in open-set domain adaptation. OVA classifiers are proposed to boost the calibration of learning to defer systems~\cite{verma_2022_calibrated}. However, the role of OVA classifiers in OOD detection has not been explored. The study of~\cite{padhy_2020_revisitingOVA} reveals that the OVA training strategy could not improve the performance of OOD detection in AUROC. In this paper, we delve into the OVA classifier and find that through suitable classifier design and training strategies, OOD detection performance can be significantly improved.

\noindent\textbf{Prototype-based methods.} Prototype-based methods represent the training set by a set of prototypical points in feature space. These prototypes could be estimated or learned from the training data. Because prototypes represent local density distributions of known classes, they can be used to reject unknown inputs. Yang \textit{et al}. proposed CPN~\cite{yang_2018_CPN,yang_2022_CPN} to model each class as a Gaussian distribution with learnable means and fixed covariance. CPN is trained using the CE loss, and a prototype loss is proposed to further enhance the model's ability to reject unknown inputs. Different from CPN, the work of~\cite{wen_discriminative_2016} assumes the covariance is a learnable diagonal matrix, which is a more general setting. Furthermore, CSSR~\cite{huang_2022_CSSR} extends the representation of class prototypes in CPN into non-linear local subspaces. They use individual auto-encoders to learn local subspace instead of a single point. A similar assumption of data distribution is also behind the Mahalanobis distance~\cite{lee_2018_maha}, which is mainly used in post-hoc techniques. Compared with previous studies, our paper mainly focuses on OVA training for prototype-based methods, which has not been explored in depth yet.

\section{Proposed One-versus-All Framework}
\label{sec:sec3}

In this section, we first introduce the background of classification tasks with rejection. Then, we elaborate on the pitfalls of training based on softmax function and cross-entropy loss. Finally, we introduce the one-versus-all learning framework.

\subsection{Background of Classification with Rejection}

Considering a supervised multi-class classification task, the input space and label space are denoted as $\mathcal{X} =\mathbb{R} ^d$ and $\mathcal{Y} =\left\{ 1,2,\cdots ,K \right\}$, where $ d $ and $ K $ are the feature dimensionality and the number of known classes, respectively. Given a labeled training dataset $\mathcal{D} _{\mathrm{train}}=\left\{ \left( \boldsymbol{x}^i,y^i \right) \right\} _{i=1}^{n}$, we can learn a feature extractor $f\left( \boldsymbol{x};\theta _0 \right)$ represented by DNN with parameters $\theta_0$. After obtaining the extracted features, a linear classifier is typically used for classification. The discriminant functions (\textit{i.e.}, logits) corresponding to each class and the posterior probabilities calculated through the softmax function are formulated as follows:
\begin{align}
	g_i\left( \boldsymbol{x} \right) &=\boldsymbol{w}_i\cdot f\left( \boldsymbol{x} \right) +b_i, \label{equ:linear_logits}
	\\
	P\left( y=i\mid\boldsymbol{x} \right) &=\frac{\exp \left( g_i(\boldsymbol{x})\right)}{\sum_{j=1}^K{\exp \left( g_j(\boldsymbol{x}) \right)}}. \label{equ:ce_posterior}
\end{align}
where $\boldsymbol{w}_i$ and $b_i$ are the weight and bias parameter for the class $i$.
To simplify and avoid notational ambiguity, we denote the computed posterior probability in Eq.~(\ref{equ:ce_posterior}) as $p_{i}^{\mathrm{CE}}\left( \boldsymbol{x} \right)$, which satisfies \textit{close-set unity} (summed up to one). At the test stage, the decision of classifier is obtained by choosing the class with maximum posterior probability. In addition to classification, pattern recognition systems also consider rejecting ambiguous or outlier/novel (OOD) patterns. Ambiguity rejection is to deny classification for those patterns that are not clear to decide the class so as to reduce the risk of misclassification. So, it is also called as misclassification rejection or detection.  Outlier rejection is to detect those patterns out of the known classes so as to avoid misclassifying them as known class. The decision of classification with rejection is made based on the following function:
\begin{align}
	c^*\left( \boldsymbol{x} \right) &=\left\{ \begin{array}{c}
		\mathrm{arg}\max  g_i\left( \boldsymbol{x} \right) , r\left( \boldsymbol{x} \right) =0\\
		\textcircled{R}\,\,\,\,  \,\,\,\,\,\,\,\,\,\,\,\,\,\,\,\,\,\,\,\,\,\,\,\,\,\,   , r\left( \boldsymbol{x} \right) =1\\
	\end{array} \right. , \label{label}
	\\
	r\left( \boldsymbol{x} \right) &=\mathbf{1} \left( \phi \left( \boldsymbol{x} \right) <\delta \right), \label{key}
\end{align}

\noindent where $ \textcircled{R} $ means the decision of rejection, $\mathbf{1}\left ( \cdot \right)$ is the indicator function, and $r\left( \boldsymbol{x} \right)$ is the rejector, whose decision is made via comparing a scoring function $\phi \left( \boldsymbol{x} \right)$ and threshold $\delta$. For example, commonly used scoring functions include MSP~\cite{hendrycks_2017_baseline} and energy-based score~\cite{liu_2020_energy}:
\begin{align}
	\phi& _{\mathrm{MSP}}\left( \boldsymbol{x} \right) =\underset{i}{\max}\,\, p_{i}^{\mathrm{CE}}\left( \boldsymbol{x} \right), \label{equ:score_msp}
	\\
	\phi& _{\mathrm{EBO}}\left( \boldsymbol{x} \right) =-\log \sum_{i=1}^K{\exp \left( g_i\left( \boldsymbol{x} \right) \right)}.  \label{equ:score_ebo}
\end{align}

Softmax function is widely used to calculate the posterior probability in classification with rejection. According to the rejection rule of Chow~\cite{chow_1970_optimum}, thresholding the maximum posterior probability results in minimum risk of misclassification and rejection in closed-set classification. However, calculating posterior probabilities based on softmax under the closed-world assumption as Eq.~(\ref{equ:ce_posterior}) will raise a fundamental issue when dealing with OOD inputs: the class posterior probabilities are forced to sum up to 1, so that when inputting an outlier pattern, there is still one known class with high posterior probability, so that the outlier input is misclassified. In the scenario of open set recognition, receiving both InD and OOD inputs, it is desired that the sum of probabilities of known classes and OOD class is 1:
\begin{equation}\label{equ:open_world}
	\sum_{i=1}^K{P\left( y=i\mid\boldsymbol{x} \right)}+p_{\mathrm{OOD}}\left( \boldsymbol{x} \right) =1,
\end{equation}

\noindent so that on OOD input, the probability of OOD class is high while all the probabilities of known classes are very low. However, the estimation of OOD class probability is usually difficult because there is no or not enough OOD samples in training the model. Many previous works have tackled this issue from the viewpoint of OSR or mainly OOD detection. In this paper, we address the problem via one-versus-all learning and will show that this can achieve the goal of Eq.~(\ref{equ:open_world}).

\subsection{One-versus-All Learning Framework} 

OVA classifier treats multi-class classification as multiple binary classifiers, and then chooses the maximum posterior probability from all binary classifiers. Each binary classifier is aimed to separate one known class from the union of the rest classes, so as to be used for judge whether the input pattern belongs to the positive (known class) or not. OVA classifier has the potential of rejecting OOD inputs because the OOD class is implied in the negative class of each binary classifier, even though there is no OOD samples in training. If the OVA classifier is designed appropriately, an OOD input is desired to be rejected by all the binary classifiers (assigned low probability to all known classes).

Formally, assume we have $K$ discriminant functions $g_1\left( \boldsymbol{x} \right) ,g_2\left( \boldsymbol{x} \right) ,\cdots ,g_K\left( \boldsymbol{x} \right)$ such that every function represents a classifier for one class. For example, the discriminant functions are calculated by Eq.~(\ref{equ:ce_posterior}) in the multi-class linear classifier. Then, the sigmoid function is applied to derive the probability of classifying to class $i$ for the binary classifier:
\begin{equation}\label{equ:ova_posterior}
	p_{i}^{\mathrm{OVA}}\left( \boldsymbol{x} \right)=P\left( y=i\mid\boldsymbol{x} \right) =\sigma \left[ g_i\left( \boldsymbol{x} \right) \right].
\end{equation}

In OVA training for $ K $-class classification, the training loss is the sum of multiple binary classification tasks, each formulated as a binary cross-entropy (BCE):
\begin{equation}\label{equ:ova_loss}
	\begin{aligned}
		L_{\mathrm{OVA}}&=-\sum_{i=1}^K{y_i\log p_{i}^{\mathrm{OVA}}+\left( 1-y_i \right) \log \left( 1-p_{i}^{\mathrm{OVA}} \right)}
		\\
		&=-\left( \log p_{y}^{\mathrm{OVA}}+\sum_{i\ne y}{\log \left( 1-p_{i}^{\mathrm{OVA}} \right)} \right),
	\end{aligned}
\end{equation}
where $ y_i $ is a binary label indicating whether the input pattern belongs to class $i$ or not. In addition to the binary discriminative loss $ L_{\text{OVA}} $, a regularization term can be used to constrain the variation of parameters, such as the weight-decay ($ L_2 $) loss popularly used in multi-layer neural networks and the prototype loss used in prototype classifier~\cite{liu_2010_OVA,yang_2018_CPN}.

After OVA training, the classifier is used for classification by comparing the sigmoid probabilities or logits of $ K $ classes. More rigorously, it was shown that the $ K $ sigmoid probabilities can be transformed to $ K+1 $ posterior probabilities corresponding to $ K $ known classes and one OOD class, by using the Dempster-Shafer theory of evidence (DSTE)~\cite{liu_2005_classifier}. On obtaining $ K+1 $ posterior probabilities, the decisions of closed-set classification and OOD rejection as well as misclassification rejection can be made based on the unified rule of maximum posterior probability. 
\begin{theorem}\label{thm:ova_main}
	The multiple binary classifiers in Eq.~(\ref{equ:ova_posterior}) can be combined into a multi-class classifier. Formally, the posterior probability of the combined classifier for class $i$, denoted as $ p_{i}^{m}\left( \boldsymbol{x}\right)$, is
	\begin{equation}\label{equ:ova_posterior_combine}
		p_{i}^{m}\left( \boldsymbol{x}\right) = \frac{\exp \left( g_i(\boldsymbol{x}) \right)}{1+\sum_{j=1}^K{\exp \left( g_j(\boldsymbol{x}) \right)}}.
	\end{equation}
\end{theorem}

\noindent\textbf{\textit{Proof.}} For the binary classifier of class $ i $, the probability of an input assigned to that class is:
\begin{equation}\label{key}
	p_{i}^{b}\left( \boldsymbol{x} \right) =\sigma \left[ g_i\left( \boldsymbol{x} \right) \right].
\end{equation}

\noindent Using Dempster–Shafer theory of evidence~\cite{shafer_1976_mathematical,liu_2005_classifier}, the combined evidence of class $i$ is

\begin{equation}\label{equ:ova_m_complex}
	\begin{aligned}
		p_{i}^{m}\left( \boldsymbol{x} \right) &=\frac{1}{A}p_{i}^{b}\left( \boldsymbol{x} \right) \times \underset{j\ne i}{\overset{K}{\Pi}}\left( 1-p_{j}^{b}\left( \boldsymbol{x} \right) \right)
		\\
		& =\frac{1}{A}\times\frac{p_{i}^{b}\left( \boldsymbol{x} \right)}{1-p_{i}^{b}\left( \boldsymbol{x} \right)}\underset{j=1}{\overset{K}{\Pi}}\left( 1-p_{j}^{b}\left( \boldsymbol{x} \right) \right),
	\end{aligned}
\end{equation}
where

\begin{equation}\label{equ:ova_m_normalize}
	\begin{aligned}
		A=&\sum_{i=1}^K{p_{i}^{b} \left( \boldsymbol{x} \right) \times \underset{j\ne i,j=1}{\overset{K}{\Pi}}}\left( 1-p_{j}^{b}\left( \boldsymbol{x} \right) \right) +\underset{j=1}{\overset{K}{\Pi}}\left( 1-p_{j}^{b}\left( \boldsymbol{x} \right) \right) 
		\\
		=&\left( 1+\sum_{i=1}^K{\frac{p_{i}^{b}\left( \boldsymbol{x} \right)}{1-p_{i}^{b}\left( \boldsymbol{x} \right)}} \right) \underset{j=1}{\overset{K}{\Pi}}\left( 1-p_{j}^{b}\left( \boldsymbol{x} \right) \right).
	\end{aligned}
\end{equation}

\noindent Given Eq.~(\ref{equ:ova_m_complex}) and Eq.~(\ref{equ:ova_m_normalize}), the posterior probability of the combined classifier, \textit{i.e.}, $ p_{i}^{m}\left( \boldsymbol{x}\right)$, is simplified as follows
\begin{equation}\label{equ:ova_m_simplify}
	\begin{aligned}
		p_{i}^{m}\left( \boldsymbol{x} \right) &=\frac{p_{i}^{b}\left( \boldsymbol{x} \right)}{1-p_{i}^{b}\left( \boldsymbol{x} \right)}/\left( 1+\sum_{i=1}^K{\frac{p_{i}^{b}\left( \boldsymbol{x} \right)}{1-p_{i}^{b}\left( \boldsymbol{x} \right)}} \right) 
		\\
		&= \frac{\exp \left( g_i(\boldsymbol{x}) \right)}{1+\sum_{j=1}^K{\exp \left( g_j(\boldsymbol{x}) \right)}}.
	\end{aligned}
\end{equation}
\qed

Theorem~\ref{thm:ova_main} quantitatively provides the posterior probabilities for each class in the combined multi-class classifier based on Dempster–Shafer theory of evidence. The way this combined multi-class classifier calculates posterior probabilities is quite similar to the commonly used softmax in Eq.~(\ref{equ:ce_posterior}), but it has an additional constant 1 in the denominator. Viewing the logit of OOD class as 0, the combined probabilities can be viewed as posterior probabilities of $ K+1 $ classes ($ K $ known classes plus one OOD class), which sum up to 1:
\begin{equation}\label{key}
	\sum_{i=1}^K{p_{i}^{m}\left( \boldsymbol{x} \right)} + p_{\text{OOD}} \left( \boldsymbol{x} \right) = 1,
\end{equation}
where
\begin{equation}\label{key}
	\begin{aligned}
		p_{\text{OOD}} \left( \boldsymbol{x} \right) &= 1- \sum_{i=1}^K{p_{i}^{m}\left( \boldsymbol{x} \right)}
		\\
		&=\frac{1}{1+\sum_{j=1}^K{\exp \left( g_j\left( \boldsymbol{x} \right) \right)}}.
	\end{aligned}
\end{equation}

We suppose that the posterior probabilities in Eq.~(\ref{equ:ova_posterior_combine}) for multi-class tasks derived using DSTE can also be trained using multi-class CE loss and fixing the logit of OOD class as 0. To some extent, this approach can achieve similar effects to the standard OVA loss training. Additionally, because it has a form similar to the softmax function, it does not saturate easily during training as the sigmoid function does.

\noindent\textbf{Rejection Rule}.
Based on the $ K+1 $ class probabilities obtained by DSTE, \textit{classification decision} can be made based on maximum posterior probability akin to Bayesian decision:
\begin{equation}\label{equ:dste_decision}
	c\left( \boldsymbol{x} \right) =\begin{cases}
		\mathrm{OOD},                 if\,\,p_{\mathrm{OOD}}\left( \boldsymbol{x} \right) \ge \underset{i}{\max}\,\,p_{i}^{m}\left( \boldsymbol{x} \right);   \\
		\mathrm{Misclassification},  elif\,\,\underset{i}{\max} \,\, p_{i}^{m}\left( \boldsymbol{x} \right) \le \delta;
		\\
		\underset{i}{\mathrm{arg}\max}\,\,p_{i}^{m}\left( \boldsymbol{x} \right), else.  \\
	\end{cases}
\end{equation}
\noindent This can be viewed as the extension of the decision rule of Chow~\cite{chow_1970_optimum}. It can be seen that: (1) if the maximum probability is from the OOD class, the input pattern is rejected as OOD; (2) if the maximum probability is from a known class $ k $ and exceeds a threshold, the input is classified to the class $ k $, otherwise is rejected as ambiguity/misclassification.

When evaluating rejection performance with variable thresholds for calculating the AUROC, the decision rule of Eq.~(\ref{equ:dste_decision}) needs change to give a scoring function for the degree of OOD-ness. For OOD detection and misclassification detection, $p_{\mathrm{OOD}}$ and max $p_{i}^{m}$ can be taken as their respective scoring function. For unified detection of OOD and  misclassification, we need to combine them into one \textit{unified scoring function}:
\begin{equation}\label{equ:dste_scoring}
	\phi ^{K+1}\left( \boldsymbol{x} \right) =\min  \left\{ 1-p_{\mathrm{OOD}}\left( \boldsymbol{x} \right) , \underset{i}{\max}\,\,p_{i}^{m}\left( \boldsymbol{x} \right) +\varepsilon \right\}.
\end{equation}
\noindent This scoring function in Eq.~(\ref{equ:dste_scoring}) is consistent with the decision function of Eq.~(\ref{equ:dste_decision}). If the input pattern is OOD, then the value of $p_{\mathrm{OOD}}$ is large and $1-p_{\mathrm{OOD}}$ is chosen as the final score. If the input belongs to misclassification, then $\underset{i}{\max}\,\,p_{i}^{m}\left( \boldsymbol{x} \right)$ is small and becomes the final score. A calibration parameter $\epsilon$ is introduced because $ 1-p_{\mathrm{OOD}}=\sum_{i=1}^K p^m_i> \underset{i}{\max} \, p_{i}^{m} $, calibration guarantees that $ 1-p_{\mathrm{OOD}} $ is selected as score on OOD inputs.

In our experiments, we use the decision function and scoring function in Eq.~(\ref{equ:dste_decision}) and Eq.~(\ref{equ:dste_scoring}) by default. It is noteworthy that there are also other rejection rules that could be applied, like the maximum outputs of multiple binary classifiers. We will provide a comprehensive comparison about different rejection rules in ablation studies.

\section{Hybrid Learning Strategy}
\label{sec:sec4}
The OVA classifier proposed above has certain advantages over the multi-class classifier trained with CE loss in OSR, but OVA training has potential challenges in implementation. First, one challenge is that the loss function involving sigmoid function is challenging to optimize~\cite{goodfellow_2016_DLbook}, because sigmoid function is \textit{easier to saturate} than softmax function, and causes gradient vanishing when the logit is extremely negative or extremely positive. Hence, the application of sigmoid function can sometimes harm the convergence of training. Second, OVA training may encounter imbalance problem because the number of negative samples (of multiple classes) is usually much large than the number of positive samples (of one class). Third, the binary classification (one versus the rest) does not correspond directly to multi-class classification, and the combination of binary probabilities by DSTE does not guarantee the accuracy of the multi-class probabilities because of the inaccuracy of binary probability estimation. So, OVA training may result in inferior multi-class classification performance compared to training with multi-class loss such as CE, especially when the total number of classes is large.

To overcome the demerits of OVA training while maintain its advantage in OSR, we propose a hybrid learning strategy where the cross-entropy loss calculated by the combined multi-classifier is used as regularization to assist OVA training. The objective of hybrid learning is:
\begin{align}
	L&=\beta\times L_{\mathrm{OVA}}+ \left( 1-\beta \right)\times L_{\mathrm{reg}}, \label{equ:equ_ce_loss}
	\\
	L&_{\mathrm{reg}}=-\sum_{i=1}^K{\mathbf{1}\left\{ y=i \right\} }\times\log  p_{i}^{m}\left( \boldsymbol{x} \right),\label{equ:ova_equi_loss}
\end{align}
where $\beta$ is the hyper-parameter for the regularizer of the multi-class CE loss, and $\mathbf{1}\left\{ \cdot \right\}$ is the indicator function. The training objective of Eq.~(\ref{equ:equ_ce_loss}) can be viewed as the hybrid of OVA training and multi-class training. The $ p_i^m(\boldsymbol{x}) $, representing the posterior probability of ground-truth class, can be replaced by the softmax probability $ p_i^{\mathrm{CE}} $ (in Eq.~(\ref{equ:ce_posterior})). Using either $ p_i^m(\boldsymbol{x}) $ or $ p_i^{\text{CE}}(\boldsymbol{x}) $ does not differ significantly when training with positive samples (of known classes) only. However, using $ p_i^m(\boldsymbol{x}) $, training can be easily extended to the case with outlier samples as well, though in this paper, we consider the case of training with known-class samples only.

\section{Instantiation with Convolutional Prototype Network}
\label{sec:sec5}

The performance of OSR also depends on the classifier model structure, whether it is a generative model or not. General models, such as convolutional prototype network~\cite{yang_2018_CPN}, CSSR~\cite{huang_2022_CSSR}, deep K-NN~\cite{sun2022KNN} has shown superiority in OOD detection. Among them, the CPN is simpler to be implemented and trained end-to-end. We hence instantiate the proposed OVA framework with the CPN. 

The convolutional prototype network (CPN)~\cite{yang_2018_CPN} was proposed for open set recognition. It assumes the features extracted by the deep convolutional network as a Gaussian mixture model (GMM), approximated by a number of prototypes (mean vectors) by assuming identity covariance. The prototypes are learned jointly with the convolutional network weights in discriminative learning, by minimizing CE from distance-based softmax or OVA training. 

Denote the feature vector extracted by the convolutional network as $ f(\boldsymbol{x}) $ and assume one prototype $ \boldsymbol{\mu}_i $ for each known class, the distance-based softmax probability is calculated as
\begin{equation}\label{equ:cpn_posterior_simplify}
	p_{i}^{\mathrm{CE}}\left( \boldsymbol{x} \right) =\frac{\exp \left( -\xi \left\| f\left( \boldsymbol{x} \right) -\boldsymbol{\mu }_i \right\| ^2 \right)}{\sum_{j=1}^K{\exp \left( -\xi \left\| f\left( \boldsymbol{x} \right) -\boldsymbol{\mu }_j \right\| ^2 \right)}}.
\end{equation}

\noindent And for training, the CE loss is
\begin{equation}\label{equ:CE_loss}
	L_{\mathrm{CE}}(\boldsymbol{x},y;\theta)=-\sum_{i=1}^K{\mathbf{1}\left\{ y=i \right\} \times}\log p_{i}^{\mathrm{CE}}\left( \boldsymbol{x} \right),
\end{equation}	

To encourage compact within-class sample distribution for each known class, a regularization term based on maximum likelihood, called \textit{prototype loss} (PL), is calculated as
\begin{equation}\label{equ:pl_loss}
	L_{\mathrm{PL}}\left( \boldsymbol{x},y;\theta \right) =\left\| g\left( \boldsymbol{x} \right) -\boldsymbol{\mu }_y \right\| _{2}^{2}.
\end{equation}
And the total loss function for training CPN is expressed as
\begin{equation}\label{equ:ce_pl_loss}
	L_{\mathrm{CE}}+\lambda \times L_{\mathrm{PL}},
\end{equation}
where $\lambda$ is the hyper-parameter for PL. 

To implement OVA training with CPN, the core issue is how to design the binary discriminant functions. Formally, assume we have $K$ discriminant functions $g_1\left( \boldsymbol{x} \right) ,g_2\left( \boldsymbol{x} \right) ,\cdots ,g_K\left( \boldsymbol{x} \right)$ such that every function represents a binary classifier for one class. To make the discriminant function transformable to sigmoid probability, we threshold the distance of each prototype with a learnable parameter:
\begin{equation}\label{equ:ova_discriminant_function}
	g_i\left( \boldsymbol{x} \right) =-\xi\left( \left\| f\left( \boldsymbol{x} \right) -\boldsymbol{\mu }_i \right\| ^2-\tau _i \right),
\end{equation}
where $\tau_i$ is a learnable threshold for the prototype, and $\xi$ is the temperature parameter that controls the smoothness of the output probability. $\xi$ is often set as a constant.

\begin{figure}[t]
	\centering
	\begin{subfigure}[b]{0.40\textwidth}
		\centering
		\includegraphics[width=\textwidth]{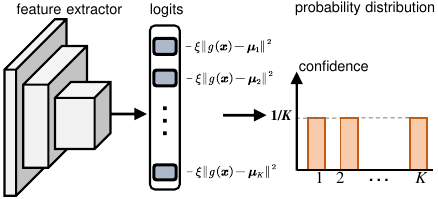}
		\caption{Multi-class classifier based on standard CE loss}
		\label{fig:img1}
	\end{subfigure}
	\hspace{0.1\textwidth}
	\begin{subfigure}[b]{0.40\textwidth}
		\centering
		\includegraphics[width=\textwidth]{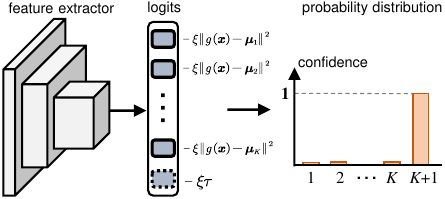}
		\caption{Combined classifier from OVA-based binary classifiers}
		\label{fig:img2}
	\end{subfigure}
	\caption{Illustration of the difference between CE-based multi-class classifier and OVA-based multi-class classifier while processing OOD inputs,. The threshold parameters play an important role in OOD detection.}
	\label{fig:OVA_ds_illustartion}
\end{figure}

The binary discriminant function of Eq.~(\ref{equ:ova_discriminant_function}) is transformed into sigmoid probability as Eq.~(\ref{equ:ova_posterior}). Then the CPN can be trained based on the OVA loss in Eq.~(\ref{equ:ova_loss}). We can also use the \textit{hybrid learning strategy} based on OVA loss, as proposed in Section~\ref{sec:sec4}, to CPN. Thus, the overall training loss function is:
\begin{equation}\label{equ:ova_pl_loss}
	\\
	\underset{\theta}{\min}\,\,\mathbb{E} _{\left( \boldsymbol{x},y \right) \sim \mathcal{D} _{\mathrm{train}}}  \beta \times L_{\mathrm{OVA}}+\left( 1-\beta \right) \times L_{\mathrm{reg}}+\lambda \times L_{\mathrm{PL}}.
\end{equation}	

\begin{algorithm*}[t]
	\caption{Hybrid learning strategy for CPN}
	\label{alg:OVA_framework}
	\renewcommand{\algorithmicrequire}{\textbf{Input:}}
	\renewcommand{\algorithmicensure}{\textbf{Output:}}
	\begin{algorithmic}[1]
		\REQUIRE Network architecture parameterized by $\theta =\left\{ \theta _0,\left( \boldsymbol{\mu }_i,\tau _i \right) _{i=1}^{K} \right\}$, training dataset $\mathcal{D} _{\mathrm{train}}$.
		\ENSURE Learned parameters $\theta$.
		\STATE Initialize network with initial weights.
		\FOR{each iteration}
		\STATE Sample a mini-batch of training samples from $\mathcal{D} _{\mathrm{train}}$.
		\STATE Calculate training loss $L_{\text{OVA}}$, $L_{\text{PL}}$, and $L_{\text{reg}}$ from Eq.~(\ref{equ:ova_loss}), Eq.~(\ref{equ:pl_loss}), and Eq.~(\ref{equ:ova_equi_loss}).
		\STATE Perform gradient descent on $\theta$ with the total loss $L_{\mathrm{total}}=\beta \times L_{\mathrm{OVA}}+\left( 1-\beta \right) \times L_{\mathrm{reg}}+\lambda \times L_{\mathrm{PL}}$.
		\ENDFOR
	\end{algorithmic}
\end{algorithm*}

\noindent We provide the pseudocode for the hybrid training framework of CPN, as shown in Algorithm~\ref{alg:OVA_framework}. 

Assuming the thresholds in Eq.~(\ref{equ:ova_discriminant_function}) for all classes are the same, CPN with OVA training can be seen as an extension of the standard CE loss classifier with the addition of an extra OOD class, as softmax-like multi-class probabilities can be derived by the DSTE, based on Theorem~\ref{thm:ova_main}. The unified threshold $ \tau $ can be viewed as the logit of the OOD class, enabling the output of $ K+1 $ probabilities. The difference between the CE-based multi-class classifier and OVA-based multi-class classifier is illustrated in Fig.~\ref{fig:OVA_ds_illustartion}.

The above learning method can also be applied to prototype classifiers with other neural network backbones, such as vision transformer (ViT)~\cite{dosovitskiy_2021_ViT}. Similarly, the backbone parameters and the prototypes are learned under the objective of OVA or hybrid learning. Considering that ViT networks pre-trained on larger datasets (such as ImageNet-1K) already have strong feature extraction capability, the strategy of learning prototypes with backbone parameters frozen can also be adopted.

\section{Experiments}
\label{sec:sec6}

We evaluate the performance of OVA learning and the hybrid learning strategy for CPN on OOD detection and MisD tasks. We verify the effectiveness of the framework on CIFAR benchmarks and ImageNet benchmark. Then, we provide comprehensive ablation studies.

\subsection{Experimental Setup}

\noindent\textbf{In-distribution Datasets.} For low-resolution datasets, we use \texttt{CIFAR-10} and \texttt{CIFAR-100}~\cite{krizhevsky2009learning} as in-distribution data. To verify the scalability of our method, we also conducted experiments using ImageNet~\cite{deng2009imagenet}. In order to improve experimental efficiency, we select the first 200 classes from ImageNet-1K~\cite{deng2009imagenet} to create \texttt{ImageNet-200}.
\begin{itemize}
	\item \texttt{CIFAR-10} consists of 60,000 32 $\times$ 32 colour images (50,000 images for training and 10,000 images for testing) in 10 classes.
	\item \texttt{CIFAR-100} consists of 60,000 32 $\times$ 32 colour images (50,000 images for training and 10,000 images for testing) in 100 classes.
	\item \texttt{ImageNet-200} consists of 267,605 224 $\times$ 224 colour images (257,605 images for training and 10,000 images for testing) in 200 classes.
\end{itemize}
\noindent\textbf{Out-of-distribution Datasets.}  For the OOD datasets in CIFAR benchmarks, we use six common benchmarks as used in previous works~\cite{liu_2020_energy,sun_2021_react,salehi_2021_OODsurvey}: \texttt{Textures}~\cite{cimpoi_2014_textures}, \texttt{SVHN}~\cite{netzer_2011_SVHN}, \texttt{LSUN-Crop}~\cite{yu_2015_lsun}, \texttt{LSUN-Resize}~\cite{yu_2015_lsun}, \texttt{Place365}~\cite{zhou_2017_places}, and \texttt{iSUN}~\cite{xu_2015_iSUN}. We would like to clarify that the CIFAR benchmarks are different from those in the earlier work~\cite{hendrycks_2017_baseline,lee_2018_maha,liang_2018_ODIN}. These new OOD datasets are harder than the earlier OOD dataset like uniform noise. As shown by recent work~\cite{liu_2020_energy,salehi_2021_OODsurvey}, most existing baselines perform far behind perfect OOD detection on this new CIFAR benchmark. For the OOD datasets in ImageNet benchmarks, we use \texttt{Textures}~\cite{cimpoi_2014_textures} as the OOD dataset. We briefly introduce the OOD datasets as below:
\begin{itemize}
	\item [$\bullet$] \texttt{Textures~\cite{cimpoi_2014_textures}} is an evolving collection of textural images in the wild, consisting of 5,640 images, organized into 47 items.
	\item [$\bullet$] \texttt{SVHN~\cite{netzer_2011_SVHN}} contains 10 classes comprised of the digits 0-9 in street view, which contains 26,032 images for testing.
	\item [$\bullet$] \texttt{Place365}~\cite{zhou_2017_places} consists in 1,803,460 large-scale photographs of scenes. Each photograph belongs to one of 365 classes. Following the work of~\cite{hendrycks_2019_oe,liu_2020_energy}, we use a subset of Place365 as OOD data for testing.
	\item [$\bullet$] \texttt{LSUN~\cite{yu_2015_lsun}} has a test set of 10,000 images of 10 different scene categories. Following~\cite{liang_2018_ODIN,liu_2020_energy}, we construct two datasets, \textit{LSUN-crop} and \textit{LSUN-resize}, by randomly cropped and downsampling LSUN test set, respectively.
	\item [$\bullet$] \texttt{iSUN~\cite{xu_2015_iSUN}} is a large-scale eye tracking dataset, selected from SUN. The dataset contains 2,000 images for the test.
\end{itemize}
\noindent The aforementioned datasets, when evaluated with models trained on CIFAR-10/100, are resized to a resolution of 32 $\times$ 32, the same size as the images in CIFAR datasets. Following common setups~\cite{hendrycks_2019_oe,liu_2020_energy,sun2022KNN}, we conduct tests by sampling 2,000 OOD samples from the dataset at a time, and repeat this process to calculate the average. When evaluated on ImageNet-200, OOD inputs are resized to 224 $\times$ 224, the same size as the images in ImageNet.

\noindent\textbf{Baseline Methods.} We compare our method with some classical and recently proposed methods in OOD detection. We mainly compare with post-hoc methods, since it is an effective line of approaches in OOD detection. The compared baselines include: MSP~\cite{hendrycks_2017_baseline}, ODIN~\cite{liang_2018_ODIN}, Mahalanobis distance (dubbed as Maha)~\cite{lee_2018_maha}, energy-based score (EBO), GradNorm~\cite{huang2021gradnorm}, ReAct~\cite{sun_2021_react}, MaxLogit~\cite{hendrycks2022MaxL}, KNN~\cite{sun2022KNN}, ASH-S~\cite{djurisic_2023_ASH}, and Neural relation graph~(Relation in brief)~\cite{kim_2023_Relation}. The hyper-parameters for these post-hoc methods come from reference code OpenOOD~\cite{Yang2022Openood} and the original work. Concretely, for ODIN~\cite{liang_2018_ODIN}, we set the temperature scaling $T$ to 1000. Specifically, for energy score, the temperature scaling $T$ is set as 1. For ReAct~\cite{sun_2021_react}, the rectification percentile is set as 90. For KNN~\cite{sun2022KNN}, the nearest neighbor number is set as 50. For ASH-S, we combine it with energy-based score as the original paper did. It is noteworthy that the shaping and scaling operations in ASH-S could affect the classification performance. For Relation, the temperature of kernel function is set to 1. Moreover, we compare our method with CPN trained by distance-based CE (DCE) loss~\cite{yang_2018_CPN,yang_2022_CPN}, to verify the advantages of the hybrid learning strategy. It is noteworthy that CPN trained by DCE loss has not been comprehensively evaluated on ODD detection and MisD for CIFAR benchmarks.

\noindent\textbf{Training Details of Baselines and Our Method.} The pre-trained models of the baselines and our model are based on \textit{almost identical configurations}. For CIFAR benchmarks, ResNet-18~\cite{he_deep_2016} and WideResNet-28-10 (WRN-28-10)~\cite{zagoruyko_wrn_2017} are the main backbones for all methods, and the model is trained by momentum optimizer with the initial learning rate of 0.1. We set the momentum to be 0.9 and the weight decay coefficient to be $2\times 10^{-4}$. Models are trained for 200 epochs and the learning rate decays with a factor of 0.1 at 100 and 150 epochs. Regarding batch size, there is a slight difference: for baseline pre-trained models, it is set as 128, while all the CPN-based models use a batch size of 64. For the pre-training of baseline models, we used the standard CE loss to train linear classifiers. For CPN, we implemented three approaches: one based on the DCE loss in Eq.~(\ref{equ:ce_pl_loss}), one based on the OVA loss and one based on the hybrid learning strategy proposed in this paper. The parameters $(\xi,\lambda,\beta)$ are set as as $ (20,0.35,0.95)$, respectively, on CIFAR-10, and as $ (2,0.05,0.95)$, respectively, on CIFAR-100.

\noindent For ImageNet benchmark, ResNet-50~\cite{he_deep_2016} is the main backbone. The configurations of pre-training models for the baselines and our method are also nearly identical. The model is trained by momentum optimizer with the initial learning rate of 0.1. The momentum is set as 0.9 and the weight decay coefficient is $1\times 10^{-4}$. Models are trained for 90 epochs and the learning rate decays with a factor of 0.1 at 30 and 60 epochs. The batch size is set as 512 for all models. For training CPN models, the parameters $(\xi,\lambda,\beta)$ are set as $ (1.5,0.05,0.8)$, respectively.

\noindent\textbf{Rejection Rule.} For the baselines, we use their post-processing-based score as the rejection rule. For CPN models, we have two approaches: for models trained with distance-based CE loss, we follow the original paper~\cite{yang_2018_CPN,yang_2022_CPN} and apply both distance-based and probability-based (dubbed as probs.-based) rejection rules. As for our method (OVA and hybrid), unless otherwise specified, we use the $ (K+1) $-class posterior probabilities derived from DSTE in Eq.~(\ref{equ:dste_scoring}) as the rejection rule (dubbed as $ (K+1) $ probs.).

\subsection{Evaluation on OOD Detection}

\noindent\textbf{Performance Metrics of OOD detection.} We use the common metrics to measure the performance of OOD detection: (1) the area under the receiver operating characteristic curve (AUROC); (2) the area under the precision-recall curve (AUPR); (3) the false positive (InD samples falsely detected as OOD) rate samples when the true positive rate of OOD samples is at 95\% (denoted as FPR95). We briefly introduce the metrics applied:
\begin{itemize}
	\item [$\bullet$] \textbf{AUROC} measures the Area Under the Receiver Operating Characteristic curve (AUROC). The ROC curve depicts the relationship between True Positive Rate and False Positive Rate.
	\item [$\bullet$] \textbf{AUPR} is the Area under the Precision-Recall (PR) curve. The PR curve is a graph showing the precision=TP/(TP+FP) versus recall=TP/(TP+FN). AUPR typically regards the in-distribution samples as positive samples.
	\item [$\bullet$] \textbf{FPR95} can be interpreted as the probability that a negative (InD data) sample is classified as an OOD input when the true positive rate is as high as 95\%.
\end{itemize}

\begin{table*}[t]
	\centering
	\caption{Performance of OOD detection on CIFAR benchmarks. All the metrics on OOD detection are the \textit{average on six OOD datasets}. ``Acc'' means the classification accuracy on in-distribution dataset. All values are percentages. The best results are boldfaced for highlight. The backbone is ResNet-18~\cite{he_deep_2016}.}
	\label{tab:main_results_res18_ood}
	\renewcommand{\arraystretch}{1.50}
	\renewcommand\tabcolsep{3.5pt}
	\scalebox{0.80}{
		\begin{tabular}{llcccccccccc}
			\specialrule{1.4pt}{0pt}{0pt}
			\multirow{2}{*}{\textbf{Model}}  &  \multirow{2}{*}{\textbf{Loss}} &  \multirow{2}{*}{\textbf{Rejection Rule}} & \multicolumn{4}{c}{In-distribution dataset: \textbf{CIFAR-10}}  &  & \multicolumn{4}{c}{In-distribution dataset: \textbf{CIFAR-100}}   \\ \cmidrule(r){4-7}  \cmidrule(r){9-12}
			& & &  \multicolumn{1}{c}{\textbf{AUROC} $\uparrow$} & \multicolumn{1}{c}{\textbf{AUPR} $\uparrow$} & \multicolumn{1}{c}{\textbf{FPR95} $\downarrow$} & \textbf{Acc} $\uparrow$ & &  \multicolumn{1}{c}{\textbf{AUROC} $\uparrow$} & \multicolumn{1}{c}{\textbf{AUPR} $\uparrow$} & \multicolumn{1}{c}{\textbf{FPR95} $\downarrow$} & \textbf{Acc} $\uparrow$  
			\\  \hline
			\multirow{8}{*}{CNN} & \multirow{8}{*}{CE} &\multicolumn{1}{l}{ODIN\textcolor{gray}{\footnotesize [ICLR2018]}~\cite{liang_2018_ODIN}}
			& \multicolumn{1}{c}{92.13}      & \multicolumn{1}{c}{97.92}     & \multicolumn{1}{c}{36.19}  &  94.98  & & \multicolumn{1}{c}{81.05}    & \multicolumn{1}{c}{95.44}   & \multicolumn{1}{c}{74.04}  & 75.83                   
			\\ 
			&  & \multicolumn{1}{l}{Maha\textcolor{gray}{\footnotesize [NeurIPS2018]}~\cite{lee_2018_maha}}
			& \multicolumn{1}{c}{90.61}      & \multicolumn{1}{c}{97.88}     & 45.31 & 94.98 & 
			& 74.36      & 92.88    & 71.97  &  75.83                         
			\\ 	
			& & \multicolumn{1}{l}{EBO\textcolor{gray}{\footnotesize [NeurIPS2020]}~\cite{liu_2020_energy}}
			& 92.24     & 97.96   & 35.15      & 94.98  &  & 81.36    & 95.50   & 72.29   &  75.83                        
			\\ 
			& &\multicolumn{1}{l}{GradNorm\textcolor{gray}{\footnotesize [NeurIPS2021]}~\cite{huang2021gradnorm}}
			& 50.04     & 81.79    & 84.18  &  94.98 & & 69.78   & 90.69    &  74.91   &  75.83                
			\\
			&&\multicolumn{1}{l}{ReAct\textcolor{gray}{\footnotesize [NeurIPS2021]}~\cite{sun_2021_react}}
			& 92.46    & 98.19   & 34.63    &  94.98  &  & 80.35    & 95.36  & 76.20  &   75.83                          
			\\ 
			&&\multicolumn{1}{l}{MaxLogit\textcolor{gray}{\footnotesize [ICML2022]}~\cite{hendrycks2022MaxL}}
			& 92.61     & 98.13  & 35.09 &  94.98 &  & 81.02    & 95.44  & 74.13    &   75.83                            
			\\ 	
			& &\multicolumn{1}{l}{KNN\textcolor{gray}{\footnotesize [ICML2022]}~\cite{sun2022KNN}}
			& \textbf{94.39}  & \textbf{98.65}   & 34.42   & 94.98  & & 80.43   & 94.84   & \textbf{69.72}   &  75.83
			\\
			& &\multicolumn{1}{l}{ASH-S\textcolor{gray}{\footnotesize [ICLR2023]}~\cite{djurisic_2023_ASH}}
			& 92.62 & 98.07 & \textbf{34.39} & \textbf{95.00} & & \textbf{82.96} &  \textbf{95.92} & 69.77  &  \textbf{76.34}
			\\
			& &\multicolumn{1}{l}{Relation\textcolor{gray}{{\footnotesize [NeurIPS2023]}}~\cite{kim_2023_Relation}}
			& 93.05 & 98.48 & 40.45 & 94.98 & & 81.62 & 95.21 & 70.54 & 75.83                  
			\\ \cdashline{3-7} \cdashline{9-12}
			&&\multicolumn{1}{l}{$K$ probs.\textcolor{gray}{\footnotesize [ICLR2017]}~\cite{hendrycks_2017_baseline} (MSP)}
			& 91.63    & 98.08    & 50.00   &  94.98 &   & 76.64   & 94.31   & 80.03  &  75.83                     
			\\
			\hline
			\multirow{4}{*}{CPN}  & \multicolumn{1}{l}{DCE+PL} & \multicolumn{1}{l}{Distance-based\textcolor{gray}{\footnotesize [T-PAMI2022]}} 
			& 93.60 & 92.09 & \textbf{33.90} & 95.08 & &  79.04 & 94.52 & 75.61 & 77.02
			\\
			& \multicolumn{1}{l}{DCE+PL} & \multicolumn{1}{l}{Probs.-based\textcolor{gray}{\footnotesize [T-PAMI2022]}}
			& 91.88 & 97.74 & 44.37 & 95.08 &  & 78.56 & 94.82 & 77.19 & 77.02
			\\
			& \multicolumn{1}{l}{OVA+PL}  &\multicolumn{1}{l}{$ (K+1) $ probs.} & 
			\cellcolor{gray!30}94.34 & \cellcolor{gray!30}98.83 & \cellcolor{gray!30}39.71 & \cellcolor{gray!30}\textbf{95.10} &   & \cellcolor{gray!30}80.89 & \cellcolor{gray!30}95.17 & \cellcolor{gray!30}72.04 & \cellcolor{gray!30}78.00
			\\
			& \multicolumn{1}{l}{hybrid+PL}  &\multicolumn{1}{l}{$ (K+1) $ probs.} & 
			\cellcolor{gray!30}\textbf{94.71} & \cellcolor{gray!30}\textbf{98.89} & \cellcolor{gray!30}36.62  & \cellcolor{gray!30}95.06 & & \cellcolor{gray!30}{\textbf{81.49}} & \cellcolor{gray!30}{\textbf{95.37}}  & \cellcolor{gray!30}{\textbf{69.94}}  & \cellcolor{gray!30}{\textbf{78.60}}   
			\\
			\specialrule{1.4pt}{0pt}{0pt}
	\end{tabular}}
\end{table*}

\begin{table*}[t]
	\centering
	\caption{Performance of OOD detection on CIFAR benchmarks. All the metrics on OOD detection are the \textit{average on six OOD datasets}. All values are percentages. The best results are boldfaced for highlight. The backbone used is WRN-28-10~\cite{zagoruyko_wrn_2017}.}
	\label{tab:main_results_wrn_ood}
	\renewcommand{\arraystretch}{1.50}
	\renewcommand\tabcolsep{3.5pt}
	\scalebox{0.80}{
		\begin{tabular}{llcccccccccc}
			\specialrule{1.4pt}{0pt}{0pt}
			\multirow{2}{*}{\textbf{Model}}  &  \multirow{2}{*}{\textbf{Loss}} &  \multirow{2}{*}{\textbf{Rejection Rule}} & \multicolumn{4}{c}{In-distribution dataset: \textbf{CIFAR-10}}  &  & \multicolumn{4}{c}{In-distribution dataset: \textbf{CIFAR-100}}   \\ \cmidrule(r){4-7}  \cmidrule(r){9-12}
			& & &  \multicolumn{1}{c}{\textbf{AUROC} $\uparrow$} & \multicolumn{1}{c}{\textbf{AUPR} $\uparrow$} & \multicolumn{1}{c}{\textbf{FPR95} $\downarrow$} & \textbf{Acc} $\uparrow$ & &  \multicolumn{1}{c}{\textbf{AUROC} $\uparrow$} & \multicolumn{1}{c}{\textbf{AUPR} $\uparrow$} & \multicolumn{1}{c}{\textbf{FPR95} $\downarrow$} & \textbf{Acc} $\uparrow$  
			\\  \hline
			\multirow{8}{*}{CNN} & \multirow{8}{*}{CE} &\multicolumn{1}{l}{ODIN\textcolor{gray}{\footnotesize [ICLR2018]}~\cite{liang_2018_ODIN}}
			& 92.35      & 97.78   & 30.63   &  95.70 &  & 81.25  & 95.42    & 74.28   &   \textbf{79.72}     \\ 
			&  & \multicolumn{1}{l}{Maha\textcolor{gray}{\footnotesize [NeurIPS2018]}~\cite{lee_2018_maha}}
			& 94.30      & 98.75    & 28.82 & 95.70   & & 63.32     & 88.83    & 79.36     &  \textbf{79.72}   \\ 	
			& & \multicolumn{1}{l}{EBO\textcolor{gray}{\footnotesize [NeurIPS2020]}~\cite{liu_2020_energy}}
			& 92.49    & 97.82     & 29.72  &  95.70 & & 78.03    & 94.47    & 78.65  &  \textbf{79.72}  \\ 
			& &\multicolumn{1}{l}{GradNorm\textcolor{gray}{\footnotesize [NeurIPS2021]}~\cite{huang2021gradnorm}}
			& \multicolumn{1}{c}{47.70}      & \multicolumn{1}{c}{85.35}     & \multicolumn{1}{c}{87.45}   &  95.70  &  & \multicolumn{1}{c}{64.90}      & \multicolumn{1}{c}{89.18}     & \multicolumn{1}{c}{81.82}      &  \textbf{79.72}   \\ 
			&&\multicolumn{1}{l}{ReAct\textcolor{gray}{\footnotesize [NeurIPS2021]}~\cite{sun_2021_react}}
			& 82.80    & 95.93    & \multicolumn{1}{c}{69.93}   &  95.70  &  & 84.09    &  \multicolumn{1}{c}{\textbf{96.33}}     & \multicolumn{1}{c}{68.46}      &  \textbf{79.72}      \\
			&&\multicolumn{1}{l}{MaxLogit\textcolor{gray}{\footnotesize [ICML2022]}~\cite{hendrycks2022MaxL}}
			& \multicolumn{1}{c}{92.35}      & \multicolumn{1}{c}{97.78}     & \multicolumn{1}{c}{30.48}   &  95.70   & & \multicolumn{1}{c}{77.82}      & \multicolumn{1}{c}{94.41}     & \multicolumn{1}{c}{79.40}      & \textbf{79.72}
			\\  
			& &\multicolumn{1}{l}{KNN\textcolor{gray}{\footnotesize [ICML2022]}~\cite{sun2022KNN}}
			& \textbf{94.40}      & \textbf{98.79}     & 32.74  &  95.70  & & 80.50  & 95.03   & 69.68    &  \textbf{79.72} 
			\\
			& &\multicolumn{1}{l}{ASH-S\textcolor{gray}{{\footnotesize [ICLR2023]}}~\cite{djurisic_2023_ASH}}
			& 92.09 & 97.50 & \textbf{28.66} & \textbf{96.04} & & \textbf{85.22} &  96.24 & \textbf{54.97}  &  79.60
			\\
			& &\multicolumn{1}{l}{Relation\textcolor{gray}{{\footnotesize [NeurIPS2023]}}~\cite{kim_2023_Relation}}
			& 94.18 & 98.70 & 33.07 & 95.70 & & 82.53 & 95.42  &  69.19 & \textbf{79.72}
			\\ \cdashline{3-12}
			&&\multicolumn{1}{l}{$K$ probs.\textcolor{gray}{\footnotesize [ICLR2017]}~\cite{hendrycks_2017_baseline} (MSP)}
			& \multicolumn{1}{c}{91.24}      & \multicolumn{1}{c}{97.72}     & \multicolumn{1}{c}{46.21}  & 95.70  &  & \multicolumn{1}{c}{75.77}   & \multicolumn{1}{c}{93.92}    & \multicolumn{1}{c}{81.53}   & 79.72    \\
			\hline
			\multirow{4}{*}{CPN}  & \multicolumn{1}{l}{DCE+PL} & \multicolumn{1}{l}{Distance-based\textcolor{gray}{\footnotesize [T-PAMI2022]}}  &
			93.30 & 98.51 & 39.42 & \textbf{96.01} & & 80.06 & 94.93 & 73.06 & \textbf{80.66}
			\\
			& \multicolumn{1}{l}{DCE+PL} & \multicolumn{1}{l}{Probs.-based\textcolor{gray}{\footnotesize [T-PAMI2022]}}  &
			90.07 & 96.49 & 44.05 & \textbf{96.01} & & 79.40 & 94.70 & 75.27 & 80.66
			\\
			& \multicolumn{1}{l}{OVA+PL}  &\multicolumn{1}{l}{$ (K+1) $ probs.} & 
			\cellcolor{gray!30}93.46 & \cellcolor{gray!30}98.57 & \cellcolor{gray!30}39.60 & \cellcolor{gray!30}95.92 & & \cellcolor{gray!30}80.15 & \cellcolor{gray!30}\textbf{95.00} & \cellcolor{gray!30}74.89 & \cellcolor{gray!30}80.29
			\\
			& \multicolumn{1}{l}{hybrid+PL}  &\multicolumn{1}{l}{$ (K+1) $ probs.}
			& \cellcolor{gray!30}\textbf{93.91}  & \cellcolor{gray!30}\textbf{98.66} & \cellcolor{gray!30}\textbf{35.76}  & \cellcolor{gray!30}95.97  &  & \cellcolor{gray!30}{\textbf{80.45}} & \cellcolor{gray!30}{94.97} & \cellcolor{gray!30}{\textbf{72.38}} & \cellcolor{gray!30} {{79.84}}	
			\\
			\specialrule{1.4pt}{0pt}{0pt}
	\end{tabular}}
\end{table*}

\noindent \textbf{Performance of InD Classification.}	Before comparing the performance on OOD detection, we first examine the classification accuracy of different methods on InD datasets. The experiments are conducted mainly on two backbones: ResNet-18~\cite{he_deep_2016} (11.2M \#params) and WRN-28-10~\cite{zagoruyko_wrn_2017} (36.5M \#params). The results can be found under the \textit{``Acc''} column in Table~\ref{tab:main_results_res18_ood} and Table~\ref{tab:main_results_wrn_ood}. From these results, it is evident that models trained using the hybrid learning strategy demonstrate promising generalization, even outperforming a conventionally trained CNN. This phenomenon is worth some explanations, especially since in the original paper of CPN~\cite{yang_2018_CPN,yang_2022_CPN}, the classification performance did not stand out. 

\begin{table*}[t]
	\centering
	\caption{Performance of MisD on CIFAR benchmarks. AURC and E-AURC values are multiplied by $ 10^3 $. Other values are percentages. The classification accuracy on in-distribution dataset is the same as Table~\ref{tab:main_results_res18_ood}. The backbone used is ResNet-18~\cite{he_deep_2016}.}
	\label{tab:main_results_res_misd}
	\renewcommand{\arraystretch}{1.50}
	\renewcommand\tabcolsep{4.5pt}
	\scalebox{0.80}{
		\begin{tabular}{llcccccccccc}
			\specialrule{1.4pt}{0pt}{0pt}
			\multirow{2}{*}{\textbf{Model}}  &  \multirow{2}{*}{\textbf{Loss}} &  \multirow{2}{*}{\textbf{Rejection Rule}} & \multicolumn{4}{c}{In-distribution dataset: \textbf{CIFAR-10}}  &  & \multicolumn{4}{c}{In-distribution dataset: \textbf{CIFAR-100}}   \\ \cmidrule(r){4-7}  \cmidrule(r){9-12}
			& & &  \multicolumn{1}{c}{\textbf{\scriptsize AUROC} $\uparrow$} & \multicolumn{1}{c}{\textbf{\scriptsize FPR95} $\downarrow$} & \multicolumn{1}{c}{\textbf{\scriptsize AURC} $\downarrow$} & \textbf{\scriptsize E-AURC} $\downarrow$ & &  \multicolumn{1}{c}{\textbf{\scriptsize AUROC} $\uparrow$} & \multicolumn{1}{c}{\textbf{\scriptsize FPR95} $\downarrow$} & \multicolumn{1}{c}{\textbf{\scriptsize AURC} $\downarrow$} & \textbf{\scriptsize E-AURC} $\downarrow$  
			\\  \hline
			\multirow{8}{*}{CNN} & \multirow{8}{*}{CE} &\multicolumn{1}{l}{ODIN\textcolor{gray}{\footnotesize [ICLR2018]}~\cite{liang_2018_ODIN}}
			&   89.01 & 53.57 & 9.26 & 8.00	& &  83.25  & 58.78 & {81.27} & {50.70}
			\\ 
			&  & \multicolumn{1}{l}{Maha\textcolor{gray}{\footnotesize [NeurIPS2018]}~\cite{lee_2018_maha}}
			&  80.02  & 62.04 & 14.52 & 13.26 & & 58.53  & 85.76  & 175.48 & 144.91 
			\\ 	
			& & \multicolumn{1}{l}{EBO\textcolor{gray}{\footnotesize [NeurIPS2020]}~\cite{liu_2020_energy}}
			& 88.73 & 62.04 & 9.42 & 8.16 & & 82.51  & 59.40 & 83.36 & 52.78
			\\ 
			& &\multicolumn{1}{l}{GradNorm\textcolor{gray}{\footnotesize [NeurIPS2021]}~\cite{huang2021gradnorm}}
			&  55.84 & 94.15 & 43.20 & 41.94 && 66.62 & 84.76 & 150.02  & 119.43
			\\ 
			&&\multicolumn{1}{l}{ReAct\textcolor{gray}{\footnotesize [NeurIPS2021]}~\cite{sun_2021_react}}
			& 89.60 & 53.25 & 8.96 & 7.69 && 81.17 & 61.01 & 87.19  & 56.64 
			\\
			&&\multicolumn{1}{l}{MaxLogit\textcolor{gray}{\footnotesize [ICML2022]}~\cite{hendrycks2022MaxL}}
			& 89.01 & 53.60 & 9.27 & 6.50 && {83.25} & {58.78} & 81.28 & 50.71 
			\\ 
			& &\multicolumn{1}{l}{KNN\textcolor{gray}{\footnotesize [ICML2022]}~\cite{sun2022KNN}}
			& 93.08 & {26.22} & {5.43} & {4.17} && 81.60 & 61.51 & 86.38 & 55.81
			\\
			& &\multicolumn{1}{l}{ASH\textcolor{gray}{{\footnotesize [ICLR2023]}}~\cite{djurisic_2023_ASH}}
			& 89.20 & 51.95 & 9.13 & 7.86 &  & 83.91 &  57.47 & 78.65  &  48.15
			\\
			& &\multicolumn{1}{l}{Relation\textcolor{gray}{{\footnotesize [NeurIPS2023]}}~\cite{kim_2023_Relation}}
			& \textbf{93.44} & \textbf{25.74} & 5.58 & 4.32 &  & 87.43 & 41.49 & 82.44 & 51.23
			\\	\cdashline{3-12}		         
			&&\multicolumn{1}{l}{$K$ probs.\textcolor{gray}{\footnotesize [ICLR2017]}~\cite{hendrycks_2017_baseline} (MSP)}
			&  \textbf{93.44} & \textbf{25.74}  & \textbf{5.39} & \textbf{4.13} && \textbf{87.80} & \textbf{41.31} & \textbf{64.46} & \textbf{33.88}
			\\
			\hline
			\multirow{4}{*}{CPN}  & \multicolumn{1}{l}{DCE+PL} & 
			\multicolumn{1}{l}{Distance-based\textcolor{gray}{\footnotesize [T-PAMI2022]}}  &
			92.43 &	38.99 &	5.46 &	4.66 & & 82.43 & 63.68 & 82.23 & 53.53
			\\
			& \multicolumn{1}{l}{DCE+PL} & \multicolumn{1}{l}{Probs.-based\textcolor{gray}{\footnotesize [T-PAMI2022]}}  &
			93.07 &	30.96 &	4.46 &	3.66 & & \textbf{87.13} & \textbf{43.99} & \textbf{63.91} & \textbf{35.22}
			\\
			& \multicolumn{1}{l}{OVA+PL}  &\multicolumn{1}{l}{$ (K+1) $ probs.} &
			\cellcolor{gray!30}93.85 & \cellcolor{gray!30}22.20 & \cellcolor{gray!30}4.79 & \cellcolor{gray!30}3.56 & & \cellcolor{gray!30}86.26 & \cellcolor{gray!30}53.28 & \cellcolor{gray!30}66.55 & \cellcolor{gray!30}40.35
			\\
			& \multicolumn{1}{l}{hybrid+PL}  &\multicolumn{1}{l}{$ (K+1) $ probs.} & 
			\cellcolor{gray!30}\textbf{93.95} & \cellcolor{gray!30}\textbf{21.71}  & \cellcolor{gray!30}\textbf{4.60}  & \cellcolor{gray!30}\textbf{3.36} & &  \cellcolor{gray!30}85.39 & \cellcolor{gray!30}55.56 & \cellcolor{gray!30}66.91 & \cellcolor{gray!30}42.18   		
			\\
			\specialrule{1.4pt}{0pt}{0pt}
	\end{tabular}}
	\vskip -0.10in
\end{table*}

\begin{table*}[t]
	\centering
	\caption{Performance of MisD on CIFAR benchmarks. AURC and E-AURC values are multiplied by $ 10^3 $. Other values are percentages. The classification accuracy on in-distribution dataset is the same as Table~\ref{tab:main_results_wrn_ood}. The backbone used is WRN-28-10~\cite{zagoruyko_wrn_2017}.}
	\label{tab:main_results_wrn_misd}
	\renewcommand{\arraystretch}{1.50}
	\renewcommand\tabcolsep{4.5pt}
	\scalebox{0.80}{
		\begin{tabular}{llcccccccccc}
			\specialrule{1.4pt}{0pt}{0pt}
			\multirow{2}{*}{\textbf{Model}}  &  \multirow{2}{*}{\textbf{Loss}} &  \multirow{2}{*}{\textbf{Rejection Rule}} & \multicolumn{4}{c}{In-distribution dataset: \textbf{CIFAR-10}}  &  & \multicolumn{4}{c}{In-distribution dataset: \textbf{CIFAR-100}}   \\ \cmidrule(r){4-7}  \cmidrule(r){9-12}
			& & &  \multicolumn{1}{c}{\textbf{\scriptsize AUROC} $\uparrow$} & \multicolumn{1}{c}{\textbf{\scriptsize FPR95} $\downarrow$} & \multicolumn{1}{c}{\textbf{\scriptsize AURC} $\downarrow$} & \textbf{\scriptsize E-AURC} $\downarrow$ & &  \multicolumn{1}{c}{\textbf{\scriptsize AUROC} $\uparrow$} & \multicolumn{1}{c}{\textbf{\scriptsize FPR95} $\downarrow$} & \multicolumn{1}{c}{\textbf{\scriptsize AURC} $\downarrow$} & \textbf{\scriptsize E-AURC} $\downarrow$  
			\\  \hline
			\multirow{8}{*}{CNN} & \multirow{8}{*}{CE} &\multicolumn{1}{l}{ODIN\textcolor{gray}{\footnotesize [ICLR2018]}~\cite{liang_2018_ODIN}}
			& 87.96 & 65.01 & 8.17 & 7.38 & & 84.23 & 57.42 & 63.22 & 41.35
			\\ 
			&  & \multicolumn{1}{l}{Maha\textcolor{gray}{\footnotesize [NeurIPS2018]}~\cite{lee_2018_maha}}
			& 86.67 & 46.82 & 7.33 &  6.54 & & 73.49  & 69.02  & 91.43 & 69.56
			\\
			& & \multicolumn{1}{l}{EBO\textcolor{gray}{\footnotesize [NeurIPS2020]}~\cite{liu_2020_energy}}
			& 92.96 & 30.74 & 8.28 & 7.48  & & 83.56 & 58.04 & 64.81 & 42.94
			\\ 
			& &\multicolumn{1}{l}{GradNorm\textcolor{gray}{\footnotesize [NeurIPS2021]}~\cite{huang2021gradnorm}}
			& 39.42 & 96.17 & 50.54 & 49.74 &  & 55.54 & 91.61 & 170.43  & 148.56 
			\\
			&&\multicolumn{1}{l}{ReAct\textcolor{gray}{\footnotesize [NeurIPS2021]}~\cite{sun_2021_react}}
			& 75.63 & 71.70  & 14.58  & 13.78 & & 70.31 & 74.75 & 104.38 & 82.51
			\\
			&&\multicolumn{1}{l}{MaxLogit\textcolor{gray}{\footnotesize [ICML2022]}~\cite{hendrycks2022MaxL}}
			& 86.77 & 69.61 & 8.94 & 8.17 &  & 84.23 & 57.42 & 63.22 & 41.36
			\\
			& &\multicolumn{1}{l}{KNN\textcolor{gray}{\footnotesize [ICML2022]}~\cite{sun2022KNN}}
			& \textbf{93.48} & \textbf{26.55} & \textbf{3.83}& \textbf{3.04} & & 84.84 & 53.04 & 60.46 & 38.59
			\\
			& &\multicolumn{1}{l}{ASH-S\textcolor{gray}{{\footnotesize [ICLR2023]}}~\cite{djurisic_2023_ASH}}
			& 88.06 & 65.53 & 8.10 & 7.30 & & 79.20 & 68.76 & 80.83 & 58.43
			\\
			& &\multicolumn{1}{l}{Relation\textcolor{gray}{{\footnotesize [NeurIPS2023]}}~\cite{kim_2023_Relation}}
			& 92.96 & 30.73 & 3.88 & 3.08 & & 88.50 & \textbf{40.52}  &  58.00 & 36.14
			\\  \cdashline{3-12}
			&&\multicolumn{1}{l}{$K$ probs.\textcolor{gray}{\footnotesize [ICLR2017]}~\cite{hendrycks_2017_baseline} (MSP)}
			& 92.97 & 30.75 & 4.63 & 3.73 & & \textbf{88.50} & \textbf{40.52} & \textbf{49.45} & \textbf{27.59} 
			\\
			\hline
			\multirow{4}{*}{CPN}  & \multicolumn{1}{l}{DCE+PL} & \multicolumn{1}{l}{Distance-based\textcolor{gray}{\footnotesize [T-PAMI2022]}}  &
			93.07 & 30.96 & 4.46 & 3.66 & & 84.90 & 62.25 & 60.51 & 40.61
			\\
			& \multicolumn{1}{l}{DCE+PL} & \multicolumn{1}{l}{Probs.-based\textcolor{gray}{\footnotesize [T-PAMI2022]}}  &
			92.43 & 38.99 & 5.46 & 4.66 & & \textbf{87.96}	& \textbf{43.63} & \textbf{49.00} & \textbf{28.96}
			\\
			& \multicolumn{1}{l}{OVA+PL}  &\multicolumn{1}{l}{$ (K+1) $ probs.} &
			\cellcolor{gray!30}93.22 & \cellcolor{gray!30}29.91 & \cellcolor{gray!30}4.49  & \cellcolor{gray!30}3.65 & & \cellcolor{gray!30}86.80 & \cellcolor{gray!30}52.13 & \cellcolor{gray!30}55.79 & \cellcolor{gray!30}34.75
			\\
			& \multicolumn{1}{l}{hybrid+PL}  &\multicolumn{1}{l}{$ (K+1) $ probs.}
			& \cellcolor{gray!30}\textbf{93.30} & \cellcolor{gray!30}\textbf{28.04} & \cellcolor{gray!30}\textbf{4.39} & \cellcolor{gray!30}\textbf{3.57} &  & \cellcolor{gray!30}86.86 & \cellcolor{gray!30}53.68  & \cellcolor{gray!30}57.10  & \cellcolor{gray!30}35.26
			\\
			\specialrule{1.4pt}{0pt}{0pt}
	\end{tabular}}
\end{table*}

\begin{figure*}[t]
	\begin{center}
		\includegraphics[width=0.60\linewidth]{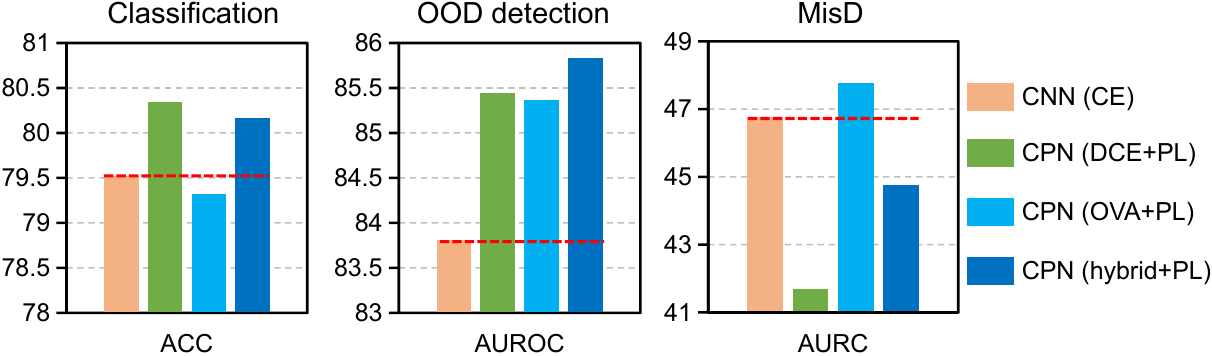}
	\end{center}
	\vskip -0.15in
	\caption{Classification with rejection on large-scale ImageNet-200/ResNet-50. Our method achieves consistent improvement on classification accuracy, MisD, and OOD detection, compared with CNN (CE). AURC is multiplied by $ 10^3 $. Other values are percentages. The rejection rule of our method is based on $ (K+1) $-class posterior probabilities. Other methods are based on $ K $-class posterior probabilities.}
	\label{fig:imagenet_exp}
	\vskip -0.10in
\end{figure*}

\noindent In some sense, CPN can be viewed as a generative classifier, which has the advantage of requiring less training data compared to discriminative classifiers, even in deep neural networks~\cite{pang_2018_mmlda,zheng_2023_revisit}.  Thus, CPN may yields higher generalized accuracy in the case of few training per class, specifically in the case of CIFAR-100 where each class has few samples than CIFAR-10. Additionally, through experiments, we found that CPN also requires a smaller batch size for training. Increasing the batch size from 64 to 128 on CIFAR-100/ReNet-18, for example, results in about 1\% decrease in accuracy. The results on ImageNet-200 as shown in Fig.~\ref{fig:imagenet_exp} indicates that CPN trained by hybrid learning yields comparably high accuracy to that trained by DCE while outperforms CPN trained by OVA learning, which degrades in the case of large number classes.

\noindent \textbf{Performance of OOD Detection.} We compare our method with \textbf{\textit{mostly recent}} methods. The results on CIFAR benchmarks are listed in Table~\ref{tab:main_results_res18_ood} and Table~\ref{tab:main_results_wrn_ood}. As can be seen, CPN trained by OVA or hybrid learning achieves superior performance than the ordinary CNN (\textit{i.e.}, MSP). For example, on CIFAR-10/CIFAR-100, CPN (hybrid+PL) achieves a remarkable gain of 3.08\%/4.85\% for ResNet-18 in terms of AUROC, compared with MSP. It is noteworthy that our rejection rule is derived from classifier output probabilities, theoretically \textit{requiring no post-hoc techniques}. Fig.\ref{fig:imagenet_exp} also shows that on ImageNet-200, all the CPN variations outperform the ordinary CNN in OOD detection, while CPN (hybrid+PL) performs best.

\noindent \textbf{Comparison with Post-hoc Techniques.} Although post-processing is not required, models trained using the OVA framework can achieve comparable or better OOD detection performance models that undergo post-processing, as shown in Table~\ref{tab:main_results_res18_ood} and Table~\ref{tab:main_results_wrn_ood}, from classifier output probabilities. KNN performs best in OOD detection among, but it \textit{can add extra overhead during testing}, because it requires storing features of a large number of training samples. In addition, most of the previous methods were designed specially for OOD detection, while our work is aimed for unified classification and rejection for both OOD and misclassification. We will show in Section 6.3 that most OOD detection methods perform inferiorly in MisD.

\noindent \textbf{Comparison of CPN Variations.} The results in Table~\ref{tab:main_results_res18_ood} and Table~\ref{tab:main_results_wrn_ood} show that CPN trained by DCE+PL performs well in OOD detection, outperforming most previous methods.

\subsection{Evaluation on MisD}

\noindent\textbf{Performance Metrics of MisD.} Commonly used metrics for MisD are the risk-coverage curve (AURC), the normalized AURC (E-AURC)~\cite{geifman2018biasreduced}, FPR95, and AUROC. The difference between AUROC and FPR95 in OOD detection and MisD is that the former regards OOD data as abnormal inputs while the latter regards misclassified examples as abnormal inputs. We give descriptions of the remaining metrics: AURC, and E-AURC.
\begin{itemize}
	\item [$\bullet$] \textbf{AURC} is the area under the risk-coverage curve, which is defined by the work of~\cite{geifman2018biasreduced}. The coverage indicates the ratio of samples whose confidence estimates are higher than some confidence threshold, and the risk is an error rate computed by using those samples. The formulations are listed as follows
	\begin{align}\label{key}
		&\mathrm{coverage}\,\,=
		\frac{\sum_{\boldsymbol{x}_i\sim D_{\mathrm{in}}}{\mathbf{1}\left( r\left( \boldsymbol{x}_i \right) =0 \right)}}{\mid \mathcal{D} _{\mathrm{in}} \mid},
		\\
		&\mathrm{risk}=
		\frac{\sum_{\boldsymbol{x}_i\sim \mathcal{D} _{\mathrm{in}}}{\mathbf{1}\left\{ \left( r\left( \boldsymbol{x}_i \right) =0 \right) \cap \left( c\left( \boldsymbol{x}_i \right) \ne y_i \right) \right\}}}{\sum_{\boldsymbol{x}_i\sim \mathcal{D} _{\mathrm{in}}}{\mathbf{1}\left( r\left( \boldsymbol{x}_i \right) =0 \right)}},
	\end{align}
	\noindent A low value of AURC means that correct and incorrect predictions are well-separable.
	\item [$\bullet$] \textbf{E-AURC}, known as excess AURC, is a normalization of AURC where we subtract the AURC of the best score function in hindsight~\cite{geifman2018biasreduced}.
\end{itemize}	

\noindent \textbf{Importance of MisD.} OOD detection and MisD are discussed together in previous studies~\cite{moon_2020_confidence,zhu_2022_rethinking}, and it is suggested by~\cite{jaeger_2023_call} that comprehensive evaluations in reliable predictions require comparing all relevant tasks from previously separated fields. In practical applications, it is unrealistic to assume that high-potential risks only result from OOD data and no risk from misclassification, so the model is expected to reject both OOD and misclassified inputs. \textit{Furthermore, the work of~\cite{jaeger_2023_call} claimed that none of the prevalent methods is able to outperform a softmax baseline (\textit{i.e.}, MSP) across a range of realistic failure sources.} We will show that the hybrid learning framework with CPN is a promising direction to tackle this problem. 

\noindent \textbf{Performance of MisD.} We evaluate the performance of the baselines and our method on MisD, and the results are listed in Table~\ref{tab:main_results_res_misd} and Table~\ref{tab:main_results_wrn_misd}. As can be seen, most of these other compared methods are much harmful for MisD, compared with CNN (MSP). It reveals that \textit{the improvements on OOD detection of most previous methods are at the cost of rejection ability on misclassified inputs.} However, our method could achieve improvement on MisD. For instance, on CIFAR-10, CPN (hybrid+PL) achieves a gain of 0.79~\textperthousand~for ResNet-18 in terms of AURC, compared with CNN (MSP). Our method achieves consistent improvement on rejecting both OOD and misclassified inputs, which is not realized by other methods.

Despite that our method could attain improvement on MisD, there are two issues that are worth discussing. One is that KNN~\cite{sun2022KNN} could also sometimes achieve better performance on MisD. KNN is a \textit{non-parametric estimation} method of data distribution in pattern recognition, and this characteristic may account for the improvement. However, KNN only outperforms MSP in easier dataset with large network, like CIFAR-10/WRN-28-10. In addition, KNN needs to \textbf{store a large number of samples or features}, which is a huge cost. Another point is that our method may perform slightly worse than MSP on MisD, though it outperforms all these other baselines. We attribute this to the complexity of modeling data in CIFAR-100. CPN assumes a single Gaussian component with equal covariance for each class, which might not capture the intricate data distribution effectively. Investigating the modeling of more intricate data distributions is a promising avenue for future research, as it has the potential to significantly improve our framework's performance.

The results on ImageNet-200 are shown in Fig.~\ref{fig:imagenet_exp}. As can be seen in the figure, our method outperforms CNN trained by CE loss. Furthermore, CPN trained by hybrid learning achieves comparable AURC to that trained by DCE while outperforms CPN trained by OVA learning, since the performance of OVA learning would degrade as the number of classes increases. 

\begin{table}[]
	\centering
	\caption{Ablation studies on PL and threshold settings on CIFAR-10. AURC and E-AURC values are multiplied by $ 10^3 $. Other values are percentages. The backbone is ResNet-18~\cite{he_deep_2016}.}
	\label{tab:ablations_component_cifar10}
	\renewcommand{\arraystretch}{1.40}
	\renewcommand\tabcolsep{4.0pt}
	\scalebox{0.88}{
		\begin{tabular}{ccccccc}
			\specialrule{1.4pt}{0pt}{0pt}
			\multicolumn{1}{c}{\multirow{2}{*}{PL}} & \multicolumn{1}{c}{\multirow{2}{*}{Thres.}} & \multicolumn{2}{c}{OOD detection} &  & \multicolumn{2}{c}{MisD}  \\ \cmidrule(r){3-4}  \cmidrule(r){6-7}
			&     & \multicolumn{1}{c}{\textbf{AUROC} $\uparrow$} & \textbf{FPR95} $\downarrow$ &  & \multicolumn{1}{c}{\textbf{AURC} $\downarrow$} & \textbf{E-ARUC} $\downarrow$ \\ \hline
			\XSolidBrush   & \ding{172}    
			&  92.48 & 42.52  &   &  7.24  &  5.74    
			\\  
			\XSolidBrush   & \ding{173}   
			& 93.76  &  41.98 &  & 4.77  & 3.60    
			\\ 
			\XSolidBrush   & \ding{174}   
			& 93.79 & 42.41 & & 5.13 & 3.94 
			\\
			\CheckmarkBold & \ding{174}  
			& 94.34 &  39.71 &  &  4.79 &  3.56      
			\\
			\specialrule{1.4pt}{0pt}{0pt}
	\end{tabular}}
	\vskip -0.10in
\end{table}

\begin{table}[]
	\centering
	\caption{Ablation studies on PL and threshold settings on CIFAR-100. AURC and E-AURC values are multiplied by $ 10^3 $. Other values are percentages. The backbone is ResNet-18~\cite{he_deep_2016}.}
	\label{tab:ablations_component_cifar100}
	\renewcommand{\arraystretch}{1.50}
	\renewcommand\tabcolsep{4.0pt}
	\scalebox{0.88}{
		\begin{tabular}{ccccccc}
			\specialrule{1.4pt}{0pt}{0pt}
			\multicolumn{1}{c}{\multirow{2}{*}{PL}} & \multicolumn{1}{c}{\multirow{2}{*}{Thres.}} & \multicolumn{2}{c}{OOD detection} &  & \multicolumn{2}{c}{MisD}  \\ \cmidrule(r){3-4}  \cmidrule(r){6-7}
			&     & \multicolumn{1}{c}{\textbf{AUROC} $\uparrow$} & \textbf{FPR95} $\downarrow$ &  & \multicolumn{1}{c}{\textbf{AURC} $\downarrow$} & \textbf{E-ARUC} $\downarrow$ \\ \hline
			\XSolidBrush   & \ding{172} 
			& 79.54 & 72.80 & & 64.10 & 38.15 
			\\  
			\XSolidBrush   & \ding{173}    
			&  78.96  & 77.42 &  &  73.87 & 46.59      
			\\ 
			\XSolidBrush   & \ding{174}   
			& 79.16 & 74.92 &  &  67.38 & 41.70    
			\\
			\CheckmarkBold & \ding{174}   
			&  80.89 & 72.04 & & 66.55 & 40.35   
			\\
			\specialrule{1.4pt}{0pt}{0pt}
	\end{tabular}}
\end{table}

\begin{table}[]
	\centering
	\caption{Ablation studies on rejection rules. AURC and E-AURC values are multiplied by $ 10^3 $. Other values are percentages. The backbone is WRN-28-10~\cite{zagoruyko_wrn_2017}.}
	\label{tab:ablations_rejection_rule}
	\renewcommand{\arraystretch}{1.50}
	\renewcommand\tabcolsep{4.0pt}
	\scalebox{0.85}{
		\begin{tabular}{llccccc}
			\specialrule{1.4pt}{0pt}{0pt}
			\multicolumn{1}{c}{\multirow{2}{*}{Dataset}} & \multicolumn{1}{c}{\multirow{2}{*}{Rej.}} & \multicolumn{2}{c}{OOD detection} &  & \multicolumn{2}{c}{MisD}  \\ \cmidrule(r){3-4}  \cmidrule(r){6-7}
			&     & \multicolumn{1}{c}{\textbf{AUROC} $\uparrow$} & \textbf{FPR95} $\downarrow$ &  & \multicolumn{1}{c}{\textbf{AURC} $\downarrow$} & \textbf{E-ARUC} $\downarrow$ \\ \hline
			\multirow{2}{*}{CIFAR-10}   & $\phi^{b}$    
			& 93.75  & 35.92 &   & 4.39  & 3.57 
			\\  
			& $\phi^{K+1}$  
			& 93.91 & 35.76 &  & 4.39 & 3.57 
			\\ 
			\multirow{2}{*}{CIFAR-100}   & $\phi^{b}$   
			& 79.89 & 73.83 & & 57.01 & 35.17
			\\
			& $\phi^{K+1}$ 
			& 80.45 & 72.38 &  & 57.10  & 35.26   
			\\
			\specialrule{1.4pt}{0pt}{0pt}
	\end{tabular}}
	\vskip -0.10in
\end{table}

\subsection{Ablation Studies}		
\noindent\textbf{Influence of PL and threshold settings.} In the original work of CPN~\cite{yang_2018_CPN,yang_2022_CPN}, PL was shown to be beneficial to classification and OOD detection performance. In this paper, we further evaluate the effects of PL with different settings of prototype thresholds (in Eq.~(\ref{equ:ova_discriminant_function})). The thresholds are learnable together with the CPN parameters. They can also be fixed values because by feature representation learning by DNN, the sample features of one class can be constrained in a hypersphere of fixed radius. As an intermediate case, the thresholds of all prototypes can share a common learnable value. Sharing a common threshold for all prototypes can help balance the discriminant functions of different classes, so as to benefit multi-class classification. In summary, we have three types of threshold settings: \ding{172}~a constant shared threshold across classes, \ding{173}~a shared learnable parameter, and \ding{174}~different learnable parameters for each class. All the experimental results are listed in Table~\ref{tab:ablations_component_cifar10} and Table~\ref{tab:ablations_component_cifar100} on CIFAR-10 and CIFAR-100, respectively. In the experiments where the threshold is set as a constant, we first obtain the learned threshold in the case \ding{173} where it is a variable. We then use this learned threshold as the constant threshold and continue to optimize the other parameters in the case \ding{173}. The results on CIFAR-10 show that when PL is not allowed, setting non-shared learnable threholds performs best, and PL further improves the performance. On CIFAR-100, however, CPN with constant shared threshold outperforms the ones with learnable threholds. This is because of the influence of imbalance for larger number of classes. In this case, using PL with learnable threshold shows evident benefit.

\begin{table*}[t]
	\centering
	\caption{Classification with rejection on larger dataset ImageNet-500 using DeiT-S~\cite{touvron_2021_deit} and ResNet-50~\cite{he_deep_2016}. Our method could archieve consistent improvement on classification accuracy and OOD detection, and a comparable performance on MisD, compared with the linear classifier trained by CE loss. The experimental results based on CNN (ResNet-50) training from scratch, showing that OVA training still has advantages in OOD detection, and \textit{DeiT-Proto exhibits significant improvement compared to traditional architectures}. AURC and E-AURC values are multiplied by $ 10^3 $. AUROC and AUPR are percentages. The best results are boldfaced for highlight. The second best results are highlighted with an underline.}
	\label{tab:exp_larget_dataset}
	\renewcommand{\arraystretch}{1.50}
	\renewcommand\tabcolsep{7.5pt}
	\scalebox{0.80}{
		\begin{tabular}{clllccccc}
			\specialrule{1.4pt}{0pt}{0pt}
			\multicolumn{1}{l}{\multirow{2}{*}{Model}} & \multicolumn{1}{l}{\multirow{2}{*}{Loss}}  & \multicolumn{1}{l}{\multirow{2}{*}{Rejection Rule}}  & \multicolumn{1}{l}{\multirow{2}{*}{Acc}}  & \multicolumn{2}{c}{OOD detection} &  & \multicolumn{2}{c}{MisD}  \\ \cmidrule(r){5-6}  \cmidrule(r){8-9}
			& & &  & \multicolumn{1}{c}{\textbf{AUROC} $\uparrow$} & \textbf{AUPR} $\uparrow$ &  & \multicolumn{1}{c}{\textbf{AURC} $\downarrow$} & \textbf{E-ARUC} $\downarrow$ \\ \hline
			\multirow{6}{*}{CNN}  & \multirow{6}{*}{CE} & \multicolumn{1}{l}{$K$ probs.\textcolor{gray}{\footnotesize [ICLR2017]}~\cite{hendrycks_2017_baseline} (MSP)}
			& \textbf{79.38} & 83.43 & 95.25 &  & \textbf{49.52} & \textbf{26.65}			
			\\
			\multirow{6}{*}{(ResNet-50)} & & \multicolumn{1}{l}{ODIN\textcolor{gray}{\footnotesize [ICLR2018]}~\cite{liang_2018_ODIN}}
			& \textbf{79.38} & \underline{87.79} & 96.38 & & 80.15 & 57.28			
			\\
			& & \multicolumn{1}{l}{EBO\textcolor{gray}{\footnotesize [NeurIPS2020]}~\cite{liu_2020_energy}}
			& \textbf{79.38} & \textbf{88.00} & \underline{96.41} &  & 84.26 & 61.40	 			
			\\
			& & \multicolumn{1}{l}{MaxLogit\textcolor{gray}{\footnotesize [ICML2022]}~\cite{hendrycks2022MaxL}}
			& \textbf{79.38} & \underline{87.79} & 96.38 & & 80.15 & 57.28		
			\\
			& & \multicolumn{1}{l}{KL-Div\textcolor{gray}{\footnotesize [ICML2022]}~\cite{hendrycks2022MaxL}}
			& \textbf{79.38} & 87.00 & \textbf{96.44} &  & \underline{69.73} & \underline{46.86}			
			\\
			& & \multicolumn{1}{l}{GEN\textcolor{gray}{\footnotesize [CVPR2023]}~\cite{liu_2023_gen}}
			& \textbf{79.38} & 87.47 & 96.39 &  & 74.44 & 51.56			
			\\ \hdashline
			\multirow{3}{*}{CPN} & DCE + PL & \multicolumn{1}{l}{$ K $ probs.\textcolor{gray}{\footnotesize [T-PAMI2022]}~\cite{yang_2022_CPN}}
			& \textbf{79.83} & 83.84 & 95.41 & & \textbf{48.38} & \textbf{26.52}			
			\\
			\multirow{3}{*}{(ResNet-50)}& OVA + PL &\multicolumn{1}{l}{$ (K+1) $ probs.}
			& \cellcolor{gray!30}79.27 & \cellcolor{gray!30}\textbf{86.81} & \cellcolor{gray!30}\textbf{96.18} & \cellcolor{gray!30} & \cellcolor{gray!30}51.52 & \cellcolor{gray!30}28.37			
			\\
			& hybrid + PL &\multicolumn{1}{l}{$ (K+1) $ probs.}
			& \cellcolor{gray!30}\underline{79.65} & \cellcolor{gray!30}\underline{85.94} & \cellcolor{gray!30}\underline{95.85} & \cellcolor{gray!30} & \cellcolor{gray!30}\underline{50.55} & \cellcolor{gray!30}\underline{28.28}		
			\\
			\specialrule{1.4pt}{0pt}{0pt}			
			\multirow{6}{*}{DeiT-S-Linear} & \multirow{6}{*}{CE} & \multicolumn{1}{l}{$K$ probs.\textcolor{gray}{\footnotesize [ICLR2017]}~\cite{hendrycks_2017_baseline} (MSP)} 
			& \textbf{84.07} & 86.38 & 96.17 & & \textbf{37.14} & \textbf{23.72}
			\\ 
			& & \multicolumn{1}{l}{ODIN\textcolor{gray}{\footnotesize [ICLR2018]}~\cite{liang_2018_ODIN}}
			& \textbf{84.07} & 86.99 & 95.85 &  & 63.72 & 50.30
			\\ 
			& & \multicolumn{1}{l}{EBO\textcolor{gray}{\footnotesize [NeurIPS2020]}~\cite{liu_2020_energy}} 
			& \textbf{84.07} & 85.62 & 95.20 & & 105.42 & 92.00
			\\ 
			& & \multicolumn{1}{l}{MaxLogit\textcolor{gray}{\footnotesize [ICML2022]}~\cite{hendrycks2022MaxL}} 
			& \textbf{84.07} & \underline{87.05} & 95.89 & & 62.72 & 49.30
			\\ 
			& & \multicolumn{1}{l}{KL-Div\textcolor{gray}{\footnotesize [ICML2022]}~\cite{hendrycks2022MaxL}} 
			& \textbf{84.07} & 86.90 & \underline{96.48} & & \underline{46.59} & \underline{33.17}
			\\ 
			& & \multicolumn{1}{l}{GEN\textcolor{gray}{\footnotesize [CVPR2023]}~\cite{liu_2023_gen}}
			& \textbf{84.07} & \textbf{89.92} & \textbf{97.18} & & 55.86 & 42.44
			\\  \hdashline
			\multirow{3}{*}{DeiT-S-Proto}   & DCE + PL & \multicolumn{1}{l}{$ K $ probs.\textcolor{gray}{\footnotesize [T-PAMI2022]}~\cite{yang_2022_CPN}}
			& \textbf{86.19} & 87.63 & 96.33 &  & \textbf{33.28} & \textbf{23.28}     
			\\ 
			& OVA + PL &\multicolumn{1}{l}{$ (K+1) $ probs.}
			& \cellcolor{gray!30}85.58 & \cellcolor{gray!30}\textbf{91.20} & \cellcolor{gray!30}\textbf{97.93} & \cellcolor{gray!30} &  \cellcolor{gray!30}69.36 & \cellcolor{gray!30}58.43   
			\\
			& hybrid + PL &\multicolumn{1}{l}{$ (K+1) $ probs.}
			& \cellcolor{gray!30}\underline{86.13} & \cellcolor{gray!30}\underline{90.74} & \cellcolor{gray!30}\underline{97.56} & \cellcolor{gray!30} & \cellcolor{gray!30}\underline{34.41} & \cellcolor{gray!30}\underline{24.31}   
			\\
			\specialrule{1.4pt}{0pt}{0pt}
	\end{tabular}}
\end{table*}

\noindent\textbf{Influence of rejection rules.} Besides the proposed decision rules in Eq.~(\ref{equ:dste_decision}) and Eq.~(\ref{equ:dste_scoring}), the outputs of multiple binary classifiers can also be used for rejection:
\begin{equation}\label{equ:ova_rule_new}
	\phi^b \left( \boldsymbol{x} \right) =\underset{i}{\max}\,\,p_{i}^{\mathrm{OVA}}\left( \boldsymbol{x} \right).
\end{equation}
We provide in Table~\ref{tab:ablations_rejection_rule} the results based on $ \phi^b\left( \boldsymbol{x} \right) $, and $ \phi^{K+1} $, on the same CPN model trained by hybrid learning strategy. The $ (K+1) $-class classifier derived in this paper are based on the probabilities from multiple binary classifiers, hence the performance of rejection based on either binary probabilities or $ (K+1) $-class probabilities does not vary significantly. Nevertheless, the performance based on $ (K+1) $-class probabilities is consistently better.

\subsection{Results on Modern Architecture and Larger Dataset}

\textbf{Background.} Recent study~\cite{fort2021_explorOOD} has shown that modern architecture like vision transformer (ViT)~\cite{dosovitskiy_2021_ViT} can significantly improve the performance on OOD detection. To demonstrate the generality and scalability of the proposed method, we conduct experiments on ViT backbone and ImageNet-subset with larger number of classes. Concretely, the backbone used in this paper is DeiT-S~\cite{touvron_2021_deit}, and the training set is \texttt{ImageNet-500}, consisting of the first 500 classes from ImageNet-1K. \textit{It is noteworthy that such experimental setups are much different from the settings in Fig.~\ref{fig:imagenet_exp}, where we use ImageNet-200/ResNet-50, and these new settings pose a harder task.} Considering the differences between ViT and CNN, we refer to the ViT equipped with the prototype classifier in this section as ViT-Proto.

\noindent \textbf{Comparison.} We adopt the released pre-trained model on ImageNet-1K from the original paper of DeiT, and only train new classifiers (prototypes) while keeping other parameters frozen. Our comparison methods include three classical OOD detection methods, MSP~\cite{hendrycks_2017_baseline}, ODIN~\cite{liang_2018_ODIN}, EBO~\cite{liu_2020_energy}, as well as three recently proposed methods for large-scale scenarios, MaxLogit~\cite{hendrycks2022MaxL}, KL Divergence (KL-Div in brief)~\cite{hendrycks2022MaxL}, and Generalized ENtropy score (GEN in brief)~\cite{liu_2023_gen}. The results for ImageNet-500/DeiT-S are shown in Table~\ref{tab:exp_larget_dataset}. It reveals that the OVA-based training methods (OVA, hybrid) can outperform linear classifier trained by CE loss (\textit{i.e.}, DeiT-S-Linear) in OOD detection. Moreover, compared with OVA training method, the hybrid learning strategy can maintain the performance on OOD detection while improving the performance on MisD, which is consistent with the other experiments. Such results demonstrate that our method can be applied to ViT backbone and larger dataset. Besides that, the results based on ResNet-50 are also provided in Table~\ref{tab:exp_larget_dataset}. The effectiveness of OVA-based training methods is also verified. Although some post-processing methods may outperform the method proposed in this paper in OOD detection, they greatly harm the model's ability to reject misclassified examples, such as EBO. Furthermore, we noticed that when applying the proposed method to ViT backbone, the improvement is more pronounced. It outperforms all compared recent OOD detection methods while also exhibiting strong misclassification detection performance. This indicates that the proposed method could be more suitable for use with ViT backbone in large-scale scenarios. Pre-training ViT on large-scale datasets and then fine-tuning the classifier using the method proposed in this paper presents a good unified performance for classification and rejection. Therefore, we will detail in subsequent part how to fine-tune the classifier using the proposed strategy.

\noindent \textbf{Training guidelines.} We would like to clarify more about the training details, to offer guidance on training prototype classifiers based on ViT. Here is the pipeline: (1)~Initializing the feature extractor using a network pre-trained on a large dataset; (2)~Initializing the prototype classifier using the mean and variance estimated from deep feature representations; (3)~Training the prototype classifier using the downstream dataset. In our experiments, we find that a proper initialization plays an essential role in training ViT-Proto, which could be estimated by following expressions:
\begin{align}
	\boldsymbol{\mu}_i &=\frac{1}{\mid\mathcal{D} _i \mid}\sum_{\left( \boldsymbol{x}_i,y_i \right) \in \mathcal{D} _i}{f\left( \boldsymbol{x} \right)},  
	\\
	var_i &=\frac{1}{\mid \mathcal{D} _i \mid}\sum_{\left( \boldsymbol{x}_i,y_i \right) \in \mathcal{D} _i}{\left\| f\left( \boldsymbol{x} \right) -\boldsymbol{\mu }_i \right\| ^2}.
\end{align}
\noindent The initialization of means (\textit{i.e.}, $\boldsymbol{\mu}_i$) is realized by the expression above. The initialization of thresholds (\textit{i.e.}, $\tau _i$) is set to $2var_i$, which empirically achieves a better performance. The model is trained by AdamW~\cite{loshchilov_2019_AdamW} optimizer with the initial learning rate of $5\times 10^{-4}$. Models are trained for 30 epochs and the learning rate decays with a cosine annealed learning rate and learning rate warmup. Regarding batch size, it is set as 2048. All experiments were conducted on the GPU processor A6000.

\section{Conclusion and Future Work}
\label{sec:sec7}

In this paper, we attempt to build a unified framework for building open set classifiers for multi-class classification and rejection of OOD and misclassification inputs. We formulate the open set recognition of $ K $-known-class as a $ (K+1) $-class classification problem with model trained on known-class samples only. The classifier is trained by OVA learning and the binary probabilities of $ K $ known classes can be combined into $ (K+1) $-class posterior probabilities, which enables classification and OOD/misclassification rejection in a unified framework. We further propose a hybrid learning strategy,  combining the advantages of OVA learning and multi-class discrimination. The proposed learning strategies are implemented on convolutional prototype network and prototype classifier with ViT backbone. Experiments on OOD detection and MisD demonstrate that the proposed framework achieves competitive performance in closed-set classification, OOD detection and MisD.

The proposed framework and learning strategies can also be implemented with other classifier architectures, such as autoencoder-based model and CSSR. Though we do not use OOD samples in training in our experiments, the proposed OVA and hybrid learning strategies are easily applied to the case of training with OOD samples. Our experiments evaluated the performance of OOD detection and MisD separately, though the proposed rejection rule allows both rejection types simultaneously. However, the measurement of performance of joint classification and OOD as well as MisD remains a future work.

\section*{Acknowledgments}

This work has been supported by the National Key Research and Development Program, China (No. 2018AAA0100400), National Natural Science Foundation of China (Nos.U20A20223, 62222609 and 62076236).

\bibliographystyle{plainnat}
\bibliography{reference}

\end{document}